%% file: acl_latex.tex
\pdfoutput=1

\documentclass[11pt]{article}

\usepackage[final]{acl}

\usepackage{times}
\usepackage{latexsym}
\usepackage{pifont}
\usepackage[table]{xcolor}

\usepackage[T1]{fontenc}

\usepackage[utf8]{inputenc}

\usepackage{microtype}
\usepackage{enumitem}\usepackage[utf8]{inputenc}

\usepackage{inconsolata}

\usepackage{graphicx}

\title{Toward Machine Translation Literacy: \\ How Lay Users Perceive and Rely on Imperfect Translations}

\author{
\textbf{Yimin Xiao}\textsuperscript{\ding{68}} \hspace{0.3cm} 
\textbf{Yongle Zhang}\textsuperscript{\ding{68}} \hspace{0.3cm} 
\textbf{Dayeon Ki}\textsuperscript{\ding{83}} \hspace{0.3cm} 
\textbf{Calvin Bao}\textsuperscript{\ding{83}} \hspace{0.3cm} 
\textbf{Marianna J. Martindale}\textsuperscript{\ding{68}} \\[0.1cm]   
\textbf{Charlotte Vaughn}\textsuperscript{\ding{118}} \hspace{0.3cm} 
\textbf{Ge Gao}\textsuperscript{\ding{68}} \hspace{0.3cm} 
\textbf{Marine Carpuat}\textsuperscript{\ding{83}} \\[0.2cm]  
\textsuperscript{\ding{68}}College of Information \hspace{0.2cm}
\textsuperscript{\ding{83}}Department of Computer Science \hspace{0.2cm} 
\textsuperscript{\ding{118}}Language Science Center \\
University of Maryland, College Park \\
\texttt{yxiao@umd.edu}
}

\usepackage{graphicx} 
\usepackage{booktabs} 
\usepackage{array} 
\usepackage{multirow} 
\usepackage{xspace}
\usepackage{longtable}
\usepackage{enumitem}

\input{ourterminology}

\begin{document}
\maketitle
\begin{abstract}
As Machine Translation (MT) becomes increasingly commonplace, understanding how the general public perceives and relies on imperfect MT is crucial for contextualizing MT research in real-world applications. We present a human study conducted in a public museum ($n=452$), investigating how fluency and adequacy errors impact bilingual and non-bilingual users' reliance on MT during casual use. Our findings reveal that non-bilingual users often over-rely on MT due to a lack of evaluation strategies and alternatives, while experiencing the impact of errors can prompt users to reassess future reliance. This highlights the need for MT evaluation and NLP explanation techniques to promote not only MT quality, but also MT literacy among its users.
\end{abstract}


\input{latex/00_intro}

\input{latex/01_background}
\input{latex/02_methods}

\input{latex/03_findings}

\input{latex/05_conclusion}

\section*{Limitations}

Our research has several limitations.

Conducting the study in a museum setting introduced several constraints. The low-stakes nature of the task, featuring a fictional interlocutor and minimal consequences for incorrect decisions, may not fully capture real-world decision-making dynamics. Additionally, the short interaction duration limits the ability to examine trust development over time \citep{holliday-wilson-stumpf-2016-usertrust}.

Our participant pool also reflects a selection bias; while it likely includes a broader age range than typical university lab studies, museum visitors who choose to engage with a scientific study are not necessarily representative of the general population and may scrutinize translations more carefully than typical users.

Furthermore, our study focuses on a single decision-making task closely aligned with reading comprehension, which may not capture the complexity of real-world MT usage scenarios.

Therefore, we caution against overgeneralizing our findings and highlight the need for further research to explore the relationships between MT quality perception, decision-making, and trust across users with varying language proficiency levels and in diverse real-world scenarios. 

\input{latex/06_acknowledgements}


\bibliography{custom}

\onecolumn
\appendix

\section{Appendix}
\label{sec:appendix}

\subsection{Full Task of an Example Stimulus}\label{app:full_task}

\input{figs/appendix_fulltask}

\newpage

\subsection{Final Stimuli}
\label{app:final_stimuli}

\input{figs/appendix_fullstimuli}

\end{document}

%% file: ourterminology.tex
\newcommand{\mt}{\textsc{MT}\xspace}

\newcommand{\correct}{Correct\xspace}
\newcommand{\incorrect}{Incorrect\xspace}
\newcommand{\fluencyerrorno}{\textsc{Fluency Error without Impact}\xspace}
\newcommand{\adequacyerrorno}{\textsc{Adequacy Error without Impact}\xspace}
\newcommand{\adequacyerroryes}{\textsc{Adequacy Error with Impact}\xspace}

\newcommand{\believability}{Perception of Translation Quality\xspace}
\newcommand{\accuracy}{Decision-Making Accuracy\xspace}
\newcommand{\confidence}{Decision-Making Confidence\xspace}
\newcommand{\trust}{Willingness to Reuse\xspace}

\newcommand{\compstrategy}{Comparison Strategy\xspace}
\newcommand{\profstrategy}{Proficiency Strategy\xspace}

\newcommand{\noprof}{No Spanish Proficiency\xspace}
\newcommand{\someprof}{Some Spanish Proficiency\xspace}
\newcommand{\highprof}{High Spanish Proficiency\xspace}

%% file: latex/00_intro.tex
\section{Introduction}

As machine translation (MT) becomes more deeply embedded in daily life through apps and chatbots, people increasingly rely on it for casual, everyday tasks: understanding signs, browsing foreign-language content, and making quick decisions. While this wide adoption signals the success of NLP technologies, it also raises questions about public understanding and appropriate use \citep{carpuat2025interdisciplinaryapproachhumancenteredmachine}. Are users equipped to detect errors or understand their consequences? Can they calibrate their trust in systems? Do they know what MT can and cannot do?  In other words, as the reach of MT has increased, what do we know about the general public’s MT literacy \citep{bowker-ciro-2019-framework}? 

This paper responds to this year’s EMNLP theme of “Advancing our Reach: Interdisciplinary Recontextualization of NLP,” which calls for rigorous evaluation of how NLP technologies actually impact society and intersect with other fields. While benchmark scores for MT continue to improve, these evaluations alone do not tell us how the general public perceives and relies on MT. Work in Translation Studies emphasizes the need for MT literacy as translation tools gain a broad range of users, who may lack the language proficiency or background knowledge to critically evaluate outputs \citep{obrien-ehrensberger-dow-2020-mt,bowker-2025-machine}. However, designing interventions that can support such a large and diverse population requires a better understanding of how people interact with MT in the wild: how they perceive its quality, how they rely on it, and  what might influence those decisions.

In this work, we study how people's reliance on MT is impacted by fluency and adequacy errors during casual, low-stakes use. Our study builds on that of \citet{martindale-carpuat-2018-fluency}, which measured user trust in presence of MT errors, but without controlling for their impact on decision-making. Here, we go further by drawing from  HCI methods for studying trust and reliance in AI \citep{vereschak-etal-2021-how}. We also conduct our experiment in a public museum setting \citep{vaughn-etal-2024-language}, enabling us to recruit participants from many walks of life and ground MT use in a specific environment.

We found that bilingual and non-bilingual users rely on MT differently, as can be expected. More surprisingly, we found that non-bilingual users often rely on imperfect MT not because they assume the outputs to be correct, but because they lack strategies to approach assessing outputs and making decisions on their basis. Interestingly, experiencing the impact of MT errors in low-stakes settings still prompted users to reevaluate their future use of the tool. 
%
These findings motivate several directions for future MT and NLP research, including the development of MT systems that support users in assessing and recovering from errors, and the development of tools to support MT literacy training inspired by the task conducted here. In the process, we hope to illustrate the benefits of recontextualizing MT and NLP work in an interdisciplinary fashion to address the societal implications of MT.

%% file: latex/01_background.tex
\section{Research Questions \& Background}
\label{sec:background}

The research questions (RQs) addressed in this paper are motivated by a body of work spanning the translation studies, HCI and MT literatures.

\paragraph{How is \mt Used?}
This is a hard question to answer because \mt is available to anyone with an internet connection \citep{savoldi-etal-2025-translation}. By 2021, the Google Translate app alone had over a billion installations \citep{pittman-2021-google}, with an estimated 99.97\% of \mt users being non-professionals \citep{nurminen-2021-investigating}. Surveys of UK residents show high satisfaction for low-stakes uses \citep{vieira-etal-2022-machine}, but public service professionals also frequently use \mt in their work without formal training \citep{nunesvieira-2024-uses}. Another concern is the use of \mt in high-stakes contexts like healthcare and law, where errors can cause significant harm \citep{khoong-etal-2019-assessing, VieiraOHaganOSullivan2021, lee-etal-2023-evaluation}. Furthermore, \mt tools do not yet meet user needs equally across socioeconomic and geographic contexts \citep{santy-etal-2021-language}, negatively impacting daily lives for groups such as migrant workers in India and immigrant populations in the U.S. \citep{liebling-etal-2020-unmet,valdez-etal-2023-migranta}.

In response, researchers emphasize the importance of raising awareness about the strengths and limitations of \mt technology \citep{VieiraOHaganOSullivan2021}. Users not only lack an understanding of how \mt operates and might fail, they also do not grasp the risks and complexities inherent in the translation process itself \citep{obrien-ehrensberger-dow-2020-mt,bowker-2025-machine}. Efforts to improve translation and \mt literacy have emerged, particularly in academic settings \citep{bowker-ciro-2019-expanding,bowker-2025-machine}. However, extending these efforts to the general public remains challenging due to the diversity of user needs and the difficulty in reaching all relevant audiences.

\paragraph{What Makes \mt``Good''?} Methods for evaluating \mt quality have evolved alongside MT technology itself.  \citet{white-etal-1993-evaluation} identified two core evaluation dimensions:  fluency, or ``well-formedness'' of the system outputs in the target language, and adequacy, ``the extent to which the semantic content of [..] texts from each source language was present in the translations''. Automatic metrics emerged to provide rapid quality assessments, comparing system outputs to professional reference translations using $n$-gram overlap \citep{papineni-etal-2002-bleu,popovic-2015-chrf} or neural methods \citep{ma-etal-2019-results, freitag2022results}. 
As MT systems advanced, human evaluation regained prominence, with protocols ranging from holistic quality ratings \citep{graham2017can}, error annotation by type and severity \citep{lommel2014multidimensional,freitag2021experts}, to post-editing of \mt outputs \citep{raunak2023leveraging, xu2023pinpoint}. 
In this approach, third party annotators evaluate translation quality to establish a ground truth rating. While this is an effective guide for system development, we also need to measure users' first person perception of \mt to address the \mt literacy gap.

\paragraph{How is MT Perceived?} 
User studies of MT and AI systems highlight that people's perception, reliance, and trust are shaped by the types of errors they encounter and their level of source language proficiency. 
Exposure to translation errors or uncertainties in AI outputs can affect users' trust and confidence in the system \cite{zhang2020effect,kim-etal-2024-im}. 
For MT, fluency errors play an important role, with evidence that they impact reported trust in MT more than adequacy errors \cite{martindale-carpuat-2018-fluency}, and that they serve as a heuristic for judging overall translation quality \cite{robertson-diaz-2022-understanding}. Further, when the outputs from MT and AI appeared to be fluent and natural sounding, people with limited proficiency experience significant challenge to understand and assess the nuanced meaning expressed in these outputs \cite{xiao-etal-2024-displaced}. 
Findings on the role of domain expertise in AI evaluation \cite{lee2023understanding, nourani2020role} also raise the question of how users’ source language proficiency influences their ability to assess translation quality and calibrate their reliance on MT.

\paragraph{Research Questions.} This contexts motivates the following Research Questions (RQs):
\begin{itemize}[leftmargin=*]
    \item[] \textbf{RQ1.} How do people perceive the quality of translations containing different error types, and subsequently, how do their decision-making accuracy and confidence vary with these error types?
    \item[] \textbf{RQ2.} How do people with varying proficiency in the source language perceive and evaluate translations containing different error types and make decisions based on these translations? 
    \item[] \textbf{RQ3.} How does people's trust in the MT system differ with exposure to different types of MT errors, and does their proficiency in the source language mediate this trust?
    \item[] \textbf{RQ4.} What primary strategies do people use in evaluating translations? And does the adopted strategy vary by people's proficiency in the source language or MT error types?
\end{itemize}

%% file: latex/02_methods.tex
\section{Methods}
\label{sec:methods}

In this section, we detail the experimental study we conducted to explore our RQs.

\subsection{Study Design}

We designed a mixed 2 $\times$ 3 experiment with two factors. \textbf{(1) MT Correctness} is \textit{within}-subject factor with two levels: \correct vs. \incorrect MT and \textbf{(2) MT Error Type} is \textit{between}-subject with three levels: \fluencyerrorno, \adequacyerrorno, and \adequacyerroryes. Participants were randomly assigned to one of three conditions.

The task is designed to mimic a low-stakes MT-mediated communication scenario in the museum setting where the study takes place \---\ thus grounding the study in a shared real-world context.
The task design is inspired by prior HCI work on MT based on controlled experiments in mock scenarios for collaboration and communication  \cite{yamashita2009difficulties,xu-etal-2014-improving,gao2013same,wang2013machine}. Prior studies have explored MT’s impact on tasks such as problem-solving \cite{yamashita2009difficulties,zhang2022facilitating} and brainstorming \cite{gao2013same,wang2013machine} in team-based settings. Others have investigated MT use in informal, everyday contexts, such as exchanging greetings, engaging in casual conversations \cite{xu-etal-2014-improving}, or facilitating housing purchases between newly arrived migrants and local people \cite{yimin_2025_sustaining}. For the museum setting, it was important to have a task that can be understood quickly, by participants of any background and any age \citep{vaughn-etal-2024-language}, and thus we exploited the only context that we know all participants share: the museum itself. Another important consideration in our task design is our focus on low-stakes scenarios that are characteristic of everyday MT use \citep{vieira-etal-2022-machine}, as these experiences are formative for users’ trust in the technology. 

\begin{figure*}[h]
    \centering
    \includegraphics[width=0.9\linewidth]{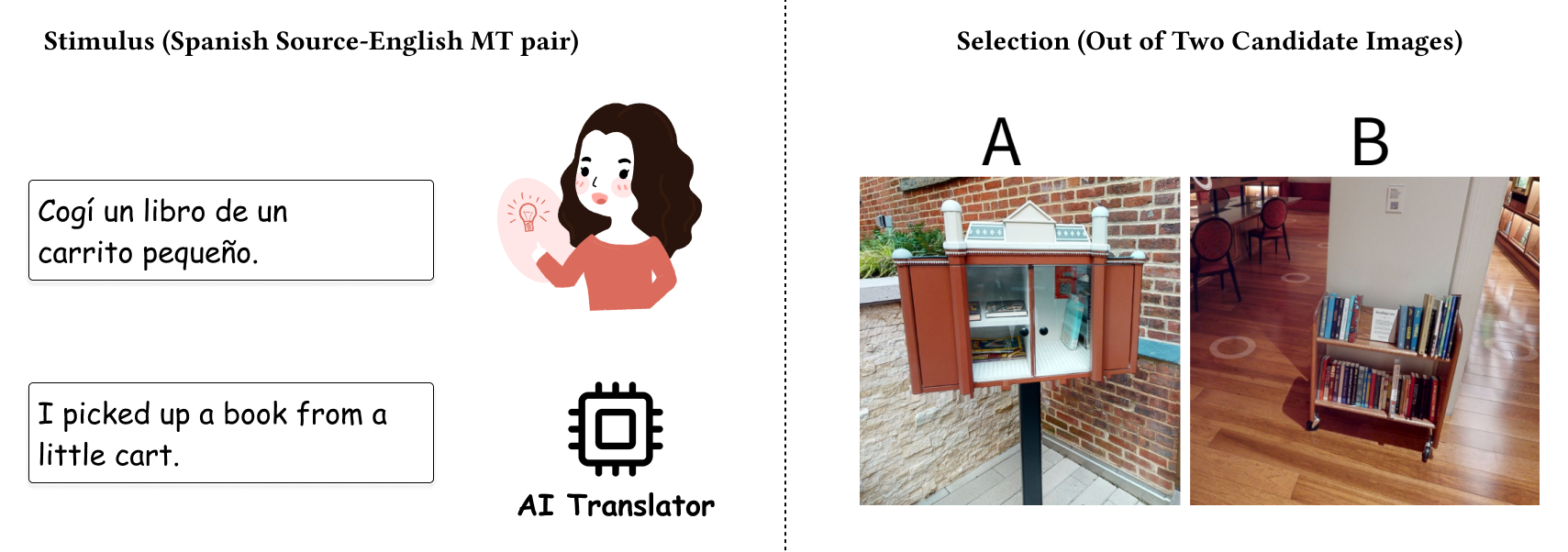}
    \caption{Example stimulus (\textbf{Stimulus}) with two candidate images for participants to select from (\textbf{Selection}). Image A shows a small outdoor library box and image B shows a small, low wooden shelf holding several books.}
    \label{fig:stimulus-example}
\end{figure*}

Participants were asked to complete a navigation task to assist a fictional Spanish-speaking character trace her steps in the museum and retrieve a lost item. The task consists of four trials (T1-T4). In each trial, participants received a stimulus composed of one Spanish message and its translated version in English. The Spanish message describes a location the fictional character was at. Upon viewing the stimulus, participants were asked to 1) rate their perception of the translation quality, 2) select one out of two images that best match the location described in the Spanish message, 3) rate their confidence in their selection. After completing each trial, participants were told whether their selection was correct or not. Figure \ref{fig:stimulus-example} shows an example stimulus with the two candidate images for participants to select from. Screenshots of the full task for an example stimulus are provided in Appendix Figure \ref{fig:appendix_fulltask}.
We manipulated the MT translation in each stimulus according to our experimental conditions. For the within-subject factor,
participants viewed correct English translations in T1 and T4 and incorrect English translations in T2 and T3. For the between-subject factors,  participants were randomly assigned into one of the three conditions (\fluencyerrorno, \adequacyerrorno, \adequacyerroryes) and viewed stimuli containing corresponding errors in T2 and T3. Presenting correct translations at the beginning (T1) helps participants calibrate their understanding of the task and establish initial trust in the translation system. Introducing errors in the middle trials (T2 and T3) enables us to observe how users respond to disruptions after forming these expectations. Within this structured ordering, we exhaustively counterbalanced the order of stimuli and randomly assigned participants to one of the pre-generated stimulus sequence. 

The experiment was implemented in Qualtrics. 
Prior to the task, participants viewed written instructions and two example stimuli. They were also to complete pre-task and post-task surveys to share demographic information, English and Spanish proficiency, and overall task experience. To ensure the clarity and engagement of our task materials, we conducted several pilots before the formal study, including think-aloud sessions across age groups and crowd-sourced surveys. Trained research assistants were available onsite to assist participants as needed, and optionally to debrief after the study for participants who wanted to know more about the study, as well as MT and generative AI.

\subsection{Stimuli Collection}

Stimuli were designed to balance study control with museum setting needs. We aimed for realistic Spanish-to-English \mt examples illustrating \correct vs. \incorrect translations, featuring three error types: \fluencyerrorno, \adequacyerrorno, and \adequacyerroryes.
Texts were kept concise and readable for broad accessibility, including for young audiences and non-native English speakers. Content was linked to the museum context to boost engagement, leveraging the only shared real-world context among visitors. Due to these constraints, using naturally occurring \mt errors was not feasible; thus, we constructed the stimuli as described below.




\paragraph{Construction.} Each stimulus includes a Spanish source sentence, its English MT, and two candidate images \---\ only one of which matches the Spanish description.
After manually creating concise and museum-relevant Spanish sentences, we generate English translations using two methods. First, we sampled LLM translations. We prompt models like \textsc{LLaMA-3} 8B \citep{grattafiori2024llama3herdmodels}, \textsc{ChatGPT}, and \textsc{GPT-4} \citep{openai2024gpt4technicalreport} with diverse decoding strategies (e.g., top-$p$, top-$k$, sampling) to produce both accurate translations and natural MT errors. Second, we introduce controlled changes to the Spanish input \citep{xu-etal-2023-understanding} \---\ such as misspellings, deletions, round-trip MT, paraphrasing, and style transfer \---\ to elicit subtle MT errors like tense shifts or word omissions.

\paragraph{Validation.}
We crowdsource three independent validation checks to ensure that text and images are interpreted as intended. These checks helped filter out ambiguous messages and unintuitive text-image pairings. Native English speakers assessed translation fluency and selected matching images based on English translations, while proficient Spanish speakers did so based on the Spanish source. Through this process, we selected 16 final stimuli across four categories (\correct, \fluencyerrorno, \adequacyerrorno, \adequacyerroryes, as illustrated in Table \ref{tab:examples}), organized into pairs for trials T1–T4. Participants were randomly assigned one T1+T4 pair and one T2+T3 pair.

\input{figs/example_table.tex}

\subsection{Participants}

We collected data at the Language Science Station (LSS; \citealt{vaughn-etal-2024-language}), a research and public engagement laboratory located in the Planet Word museum in Washington, D.C. The LSS invites museum visitors to contribute to research studies within the galleries and to engage in conversations about language with students and educators from local universities. This setting was particularly well suited to our study: it enabled us to recruit a large and diverse sample of participants with varying Spanish proficiency, experience with MT, and demographic backgrounds. Moreover, the LSS’s mission of science communication and education aligns closely with our research goal of promoting MT and AI literacy. 

Participants were required to have sufficient English proficiency to comprehend the task instruction to participate. Participants were consented prior to the start of the study and were not compensated for their participation. The study protocol was approved by the University of Maryland's Institutional Review Board.

In total, 517 people participated in our study. 65 participants did not complete the task or participated in the task with other people such as a family member or friend. After filtering out these responses, our sample included valid responses from 452 participants, including 269 females, 159 males, 15 people identified as non-binary or gender-fluid, and 9 people who preferred not to disclose their gender. Their average age was 35.12 years old (S.E. = 1.65). For Spanish proficiency, 63 participants (13.94\%) reported no proficiency at all, 353 participants (78.10\%) reported some proficiency, and 36 participants (7.96\%) reported high proficiency. For English proficiency, 406 participants (89.82\%) reported high English proficiency, with 46 participants (10.18\%) reporting some English proficiency. For self-reported usage of MT tools, 159 (35.18\%) participants never used MT, 162 participants (35.84\%) rarely used MT, 74 participants (16.37\%) sometimes used MT, 40  participants (8.85\%) often used MT, and 17 participants (3.76\%) used MT almost everyday.

\subsection{Dependent Variables}
\label{subsec:stimuli_dv}

We collected several dependent measures to address our research questions. For each stimulus, we assessed perceived translation quality, decision accuracy, and confidence. At the end of the survey, each participant reported their willingness to reuse the MT system and any evaluation strategies used.

\begin{itemize}[leftmargin=*, itemsep=2pt, parsep=-1pt]
 \item \textbf{Translation Quality Perception}: Participants rated each English translation after seeing the Spanish source, choosing whether it seemed correct (1), unclear (0.5), or problematic (0).
 \item \textbf{Decision Accuracy}: Participants selected which of two images best matched the Spanish message. A correct choice scored 1; incorrect, 0.
 \item \textbf{Decision Confidence}: Participants rated confidence in their decision on a 5-point Likert scale.
 \item \textbf{Willingness to Reuse MT}: Participants rated willingness to reuse the system on a 5-point Likert scale.
 \item \textbf{Evaluation Strategy}: Participants selected which of three strategy types best matched what they did: no strategy/intuitive, comparative analysis of Spanish and English texts, or Spanish proficiency-based judgment.
\end{itemize}

%% file: figs/example_table.tex
\begin{table*}[!htp]
\centering
\resizebox{\linewidth}{!}{%
    \begin{tabular}{p{0.3\linewidth} p{0.3\linewidth} p{0.45\linewidth} p{0.45\linewidth}}
    \toprule
    \textbf{Spanish Source} & \textbf{English MT} & \textbf{Correct Image} & \textbf{Incorrect Image} \\
    \toprule

    \rowcolor{gray!15}
    \multicolumn{4}{l}{\textbf{\fluencyerrorno}} \\
    Miré un instrumento de cuerda en una vitrina. & Look at a bristle instrument in a showcase. & 
    \parbox[l]{0.45\textwidth}{
        \includegraphics[alt={Image displaying instruments behind the glass windows of a music shop with a blue storefront labeled ``Zithers Music Shop''.}, width=0.2\textwidth]{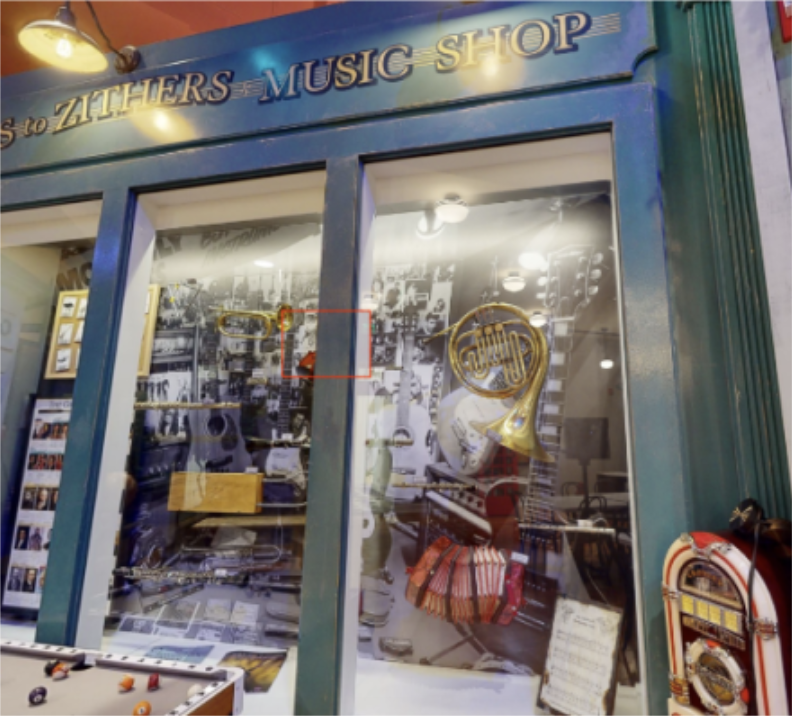} \\ 
        \footnotesize{(\textit{Instruments behind the glass windows of a music shop with a blue storefront labeled ``Zithers Music Shop''.})}
    } & 
    \parbox[l]{0.45\textwidth}{
        \includegraphics[alt={Image displaying karaoke setup on a wall, with the title ``Unlock the Music'' and visuals of lyrics and a man playing guitar.}, width=0.2\textwidth]{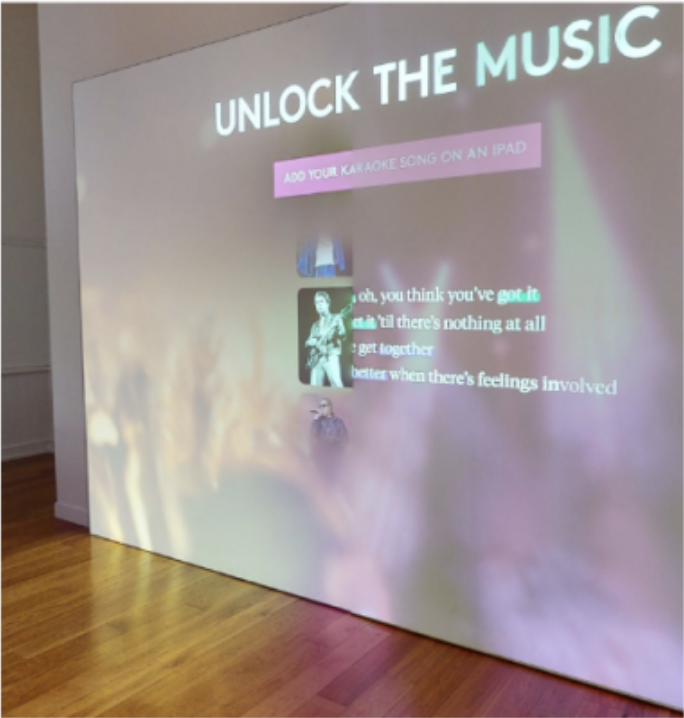} \\ 
        \footnotesize{(\textit{Karaoke setup on a wall, with the title ``Unlock the Music'' and visuals of lyrics and a man playing guitar.})}
    } \\
    \midrule

   \rowcolor{gray!15}
   \multicolumn{4}{l}{\textbf{\adequacyerroryes}} \\
    Pisé letras y señales de muchos idiomas. & Write letters and signs of many languages. & 
    \parbox[l]{0.45\textwidth}{
        \includegraphics[alt={Image displaying floor near an elevator with characters and signs in many languages spread around.}, width=0.2\textwidth]{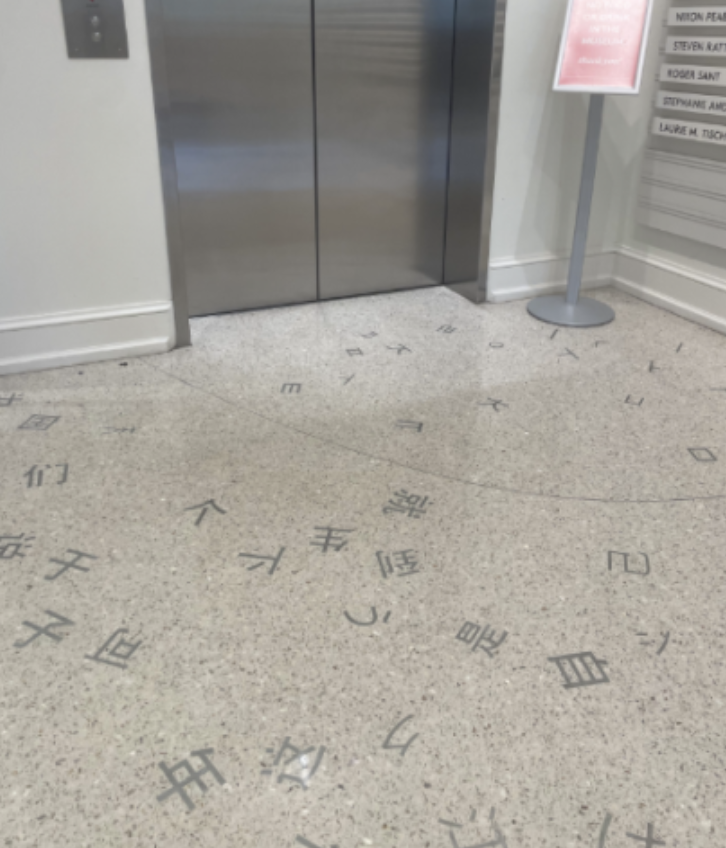} \\ 
        \footnotesize{(\textit{Floor near an elevator with characters and signs in many languages spread around.})}
    } & 
    \parbox[l]{0.45\textwidth}{
        \includegraphics[alt={Image displaying green sticky notes on a wall with the text ``One word I love from another language is ...''}, width=0.2\textwidth]{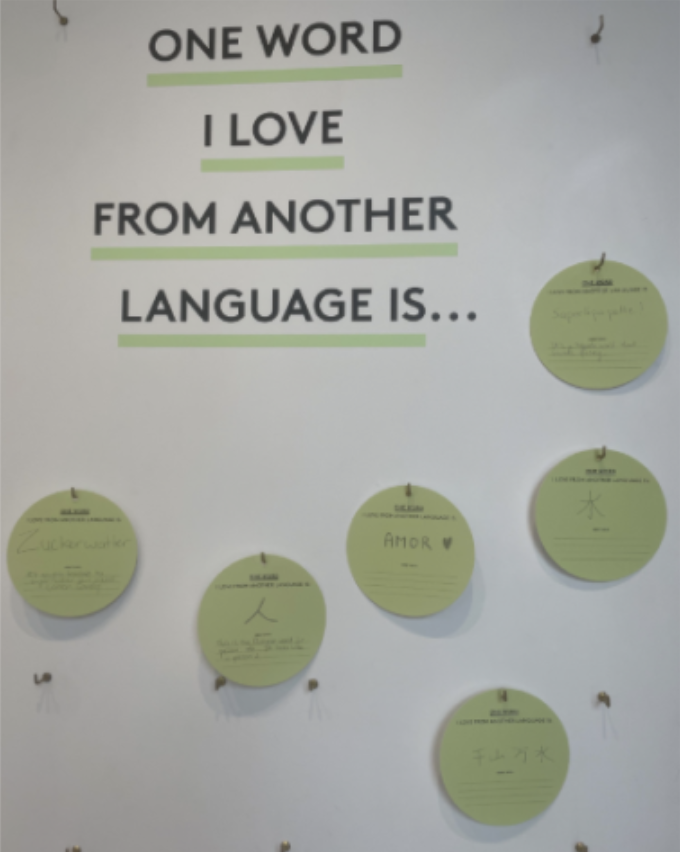} \\ 
        \footnotesize{(\textit{Green sticky notes on a wall with the text ``One word I love from another language is ...''.})}
    } \\


    \bottomrule
    \end{tabular}
}
\caption{Stimuli examples for two error types that impact the correct vs. incorrect image decision. \textbf{Spanish Source:} Spanish source sentence; \textbf{English MT:} MT of the corresponding Spanish source; \textbf{Correct Image:} Image aligned with the Spanish content; \textbf{Incorrect Image:} Image that is plausible but does not align with the Spanish. The full set of stimuli can be found in Appendix \ref{app:final_stimuli}.} 
\label{tab:examples}
\end{table*}


%% file: latex/03_findings.tex
\section{Results}
\label{sec:results}

We address each RQ in turn, by presenting the results of the statistical analysis and its implications (Section~\ref{sec:results:rq12}-\ref{sec:results:rq4}), before summarizing qualitative feedback (Section~\ref{sec:results:qualitative}). 

\subsection{RQs 1 \& 2: Perception of MT and Decision Making Based on Translations}
\label{sec:results:rq12}

The RQs 1 and 2 ask about participants' perception of MT quality and reliance on MT outputs in decision-making tasks, and how they are influenced by  MT error types and users Spanish proficiency.

\begin{figure*}[h]
    \centering
    \includegraphics[alt={Grouped bar chart titled ``Believability by MT Error Type and Spanish Proficiency''. The horizontal axis displays three groups by Spanish proficiency: No Proficiency, Some Proficiency, and High Proficiency. The vertical axis ranges from 0 to 1, representing mean believability ratings from four conditions: (1) No Error, (2) Fluency Error, (3) Adequacy Error-No Impact, and (4) Adequacy Error-Impact. Across proficiency levels, the ``No Error'' bars are highest. Significance markers (p < .05, .01, .001) indicate notable differences in believability among the error conditions. Results show that errors are statistically significant with the ''No Error' group with the error types in the ``High Proficiency'' grouping.}, width=0.9\linewidth]{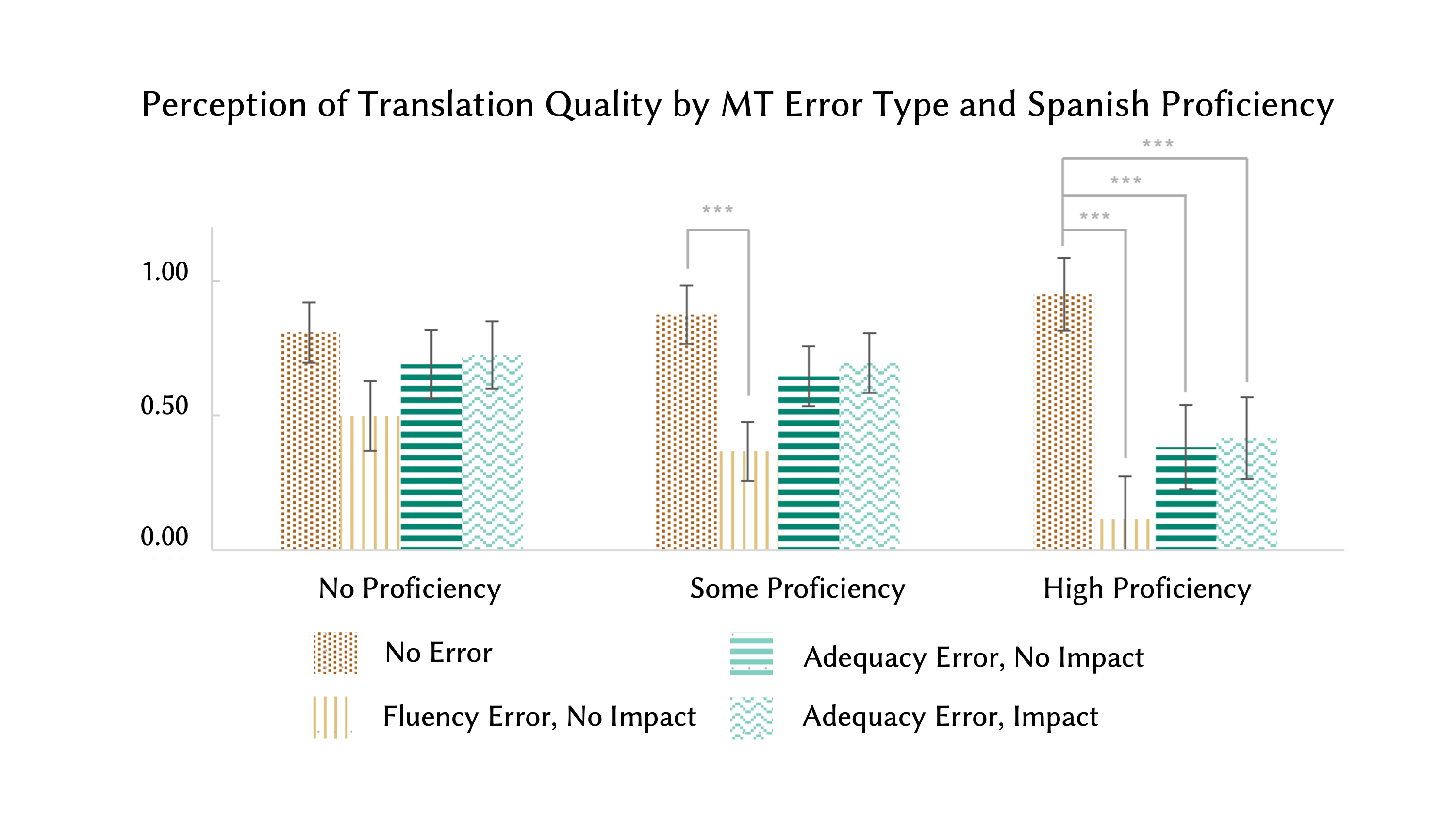}
    \caption{Perception of Translation Quality (i.e., Believability) by MT Error Type (legend) and Spanish Proficiency (x-axis). Note: *\textit{p} < .05; **\textit{p} < .01; ***\textit{p} < .001.}
    \label{fig:believability}
\end{figure*}

\paragraph{Perception of Translation Quality.} We fitted a Cumulative Link Mixed Model (CLMM) with a logit link to investigate how participants' perceptions of translation quality across error types and effects of Spanish proficiency. In the model setup, we treated participants' \believability for each stimulus as dependent variable. Our two fixed-effect independent variables were MT error types (\correct, \fluencyerrorno, \adequacyerrorno, and \adequacyerroryes) and Spanish Proficiency (\noprof, \someprof, and \highprof). We treated Participant ID and Stimulus ID as random effects. Our co-variates included individual participants' age, gender, English proficiency, and prior experience with MT. We applied Bonferroni corrections to adjust multiple comparisons.   Figure \ref{fig:believability} illustrates the core results. 

We find that participants' perception of MT quality was influenced by both their language proficiency and MT error types. \fluencyerrorno were perceived significantly less believable than \correct (Coefficient = -2.97, \textit{p} < .001). 
Participants with higher Spanish proficiency reported higher ratings of \believability  (Coefficient = 1.08, \textit{p} < .001).

Further, we observed significant interaction effects between MT error types and Spanish proficiency. Pair-wise comparisons revealed that participants with \highprof were able to detect MT errors, regardless of the error types. Participants with \someprof were able to perceive \fluencyerrorno but not \adequacyerroryes or \adequacyerrorno. Participants with \noprof were not able to perceive any MT errors, even the fluency errors which were detectable by monolingual English speakers based on our validation studies. 
%
%
%

These results shed new light on prior work suggesting that fluency and adequacy errors impact people's perceptions of MT quality differently. 
Here, participants' assessment of quality varies depending on their proficiency in the source language. Surprisingly, this is the case even when error cues are visible in the target language.




\paragraph{Decision-Making Accuracy.} We used a Generalized Linear Mixed Model (GLMM) with a logit link. Similarly, we treated MT Error Type and Spanish Proficiency as fixed-effect independent variables, and Participant ID and Stimulus ID as random effects; We applied the same set of control variables and correction method as before. 

\begin{figure*}[h]
    \centering
    \includegraphics[alt={Grouped bar chart titled ``Selection Accuracy by MT Error Type and Spanish Proficiency''. The horizontal axis displays three groups by Spanish proficiency: No Proficiency, Some Proficiency, and High Proficiency. The vertical axis ranges from 0 to 1, representing mean selection accuracy from four conditions: (1) No Error, (2) Fluency Error, (3) Adequacy Error-No Impact, and (4) Adequacy Error-Impact. 
    Generally, Adequacy Error-Impact results in the lowest bars.
    Errors are statistically significant with the ''Adequacy Error-Impact' group with the other groupings in the ``No Proficiency'' and ``Some Proficiency'' groupings.}, width=0.9\linewidth]{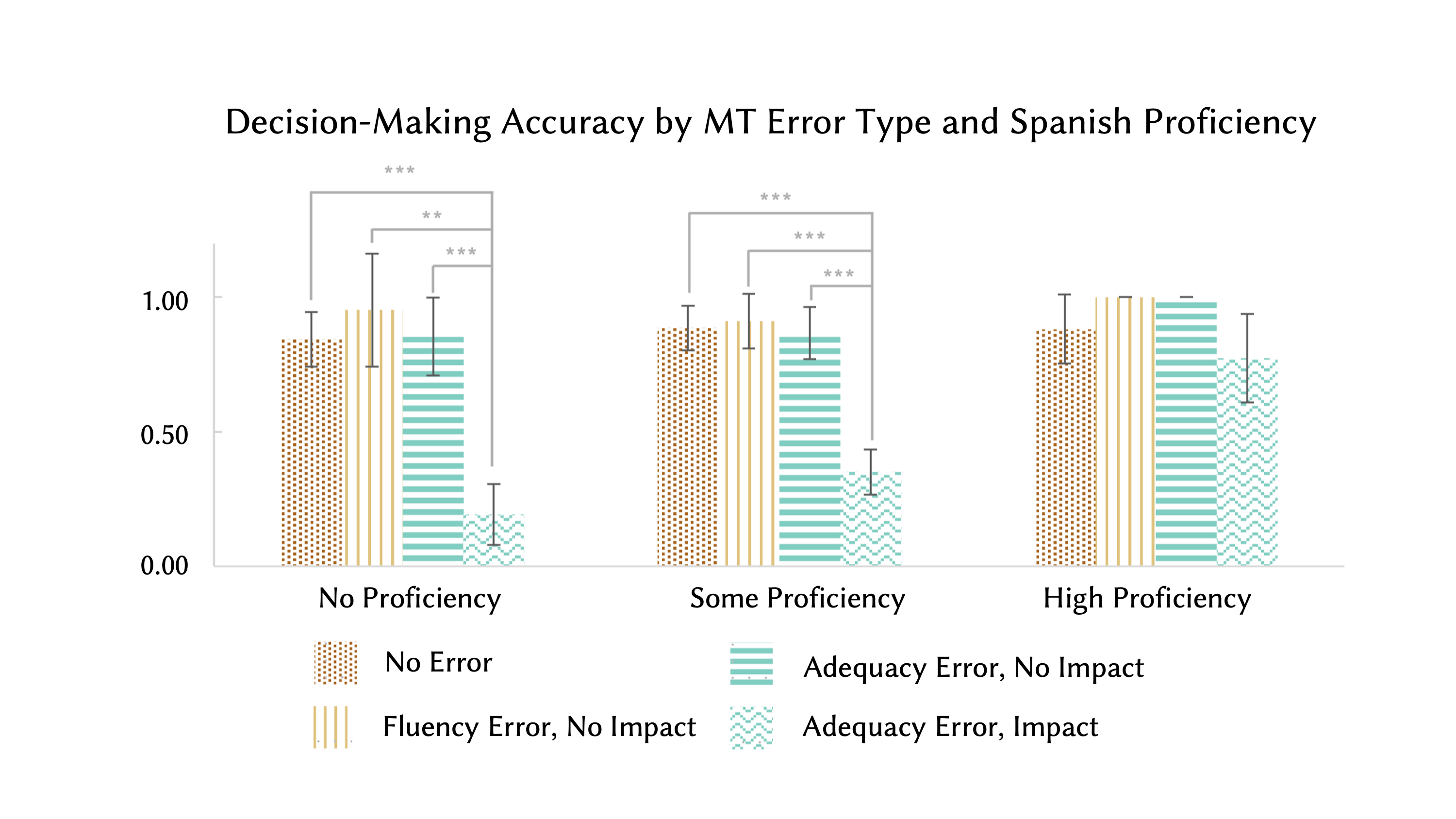}
    \caption{Decision Accuracy by MT Error Type and Spanish Proficiency. Note: *\textit{p} < .05; **\textit{p} < .01; ***\textit{p} < .001.}
    \label{fig:selection_accuracy}
\end{figure*}

The analysis shows a significant main effect for \adequacyerroryes, where participants showed significantly lower \accuracy when faced with \adequacyerroryes compared to \correct (Coefficient = -2.18; \textit{p} < .001). There was also a significant interaction between \adequacyerroryes and Spanish Proficiency, motivating pair-wise comparisons. Figure \ref{fig:selection_accuracy} illustrates the core results.

These results show that participants with No or Some Spanish Proficiency were less accurate in their \accuracy based on MT with misleading errors (\adequacyerroryes), while those with \highprof showed no difference in \accuracy across different MT errors. 
This further illustrates the disparate impact of MT errors on users depending on different proficiency levels. 

\paragraph{Decision-Making Confidence.} We used a CLMM model with a logit link, with the same set of fixed-effect, random-effect and control variables and correction method as above.
we found that Spanish Proficiency has a significant main effect on \confidence. Specifically, participants with higher Spanish proficiency demonstrated significantly higher \confidence compared to those with lower proficiency (Coefficient = 0.225, z = 2.47, \textit{p} < 0.05). However, MT error types did not show significant main effects, and there were no significant interaction effects between MT Errors types and Spanish Proficiency.

\paragraph{Recap \& Implications.} These results highlight the importance of accounting for language proficiency differences in human studies of MT and in MT interaction design. We find that bilingual and non-bilingual users rely on MT in predictably different ways, highlighting a fundamental fairness issue in the use of MT. More surprisingly, lack of Spanish proficiency impacted participants' ability to perceive fluency errors, which were in principle detectable based on the English alone, and even some knowledge of the language was not sufficient to compensate for impactful adequacy errors. 

This emphasizes the need for developing and testing MT and NLP techniques to guide error assessment. Future work could draw from methods for highlighting differences between \mt input and outputs \citep{briakou-etal-2023-explaining}, and a wealth of existing techniques to automatically estimate the quality of a translation or detect potential errors \citep{SpeciaRajTurchi2010,fomicheva-etal-2021-eval4nlp,kocmi-federmann-2023-gembamqma,guerreiro-etal-2023-xcomet,jung-etal-2024-precise}. However, how to present such information to effectively support users in practice remains an open question \citep{tsai-wang-2015-evaluating,zouhar-etal-2021-backtranslation,mehandru-etal-2023-physiciana}. 

\subsection{RQ3: Individual Participants' Willingness to Reuse MT}
\label{sec:results:rq3}

For our RQ3, we analyzed the relationship between MT Error type (three between-subject levels: \fluencyerrorno, \adequacyerrorno), and \adequacyerroryes), Spanish proficiency (three levels: \noprof, \someprof, \highprof), and participants' \trust the AI translator by evaluating their willingness to reuse the tool in the future. We analyzed results per participant, using an ANOVA to examine the main effects and interactions, and controlling for covariates such as age, gender, English proficiency, and MT experience. 

The analysis revealed a significant main effect of MT Error type on \trust in the AI translator (F(2, 374) = 4.798, \textit{p} < .01). Specifically, participants' willingness to reuse the tool varied across the error types, suggesting that the type of MT error influences users' trust in the system. To further explore these differences, we conducted a Tukey's HSD post-hoc test. Participants reported significantly lower \trust in the AI when \fluencyerrorno were present (Mean = 2.81, S.E. = 0.20) compared to conditions where there were \adequacyerrorno (Mean = 3.00, S.E. = 0.21; \textit{p} < .05). \trust was also significantly lower when participants encountered \adequacyerroryes (Mean = 2.90, S.E. = 0.20) compared to situations where there was \adequacyerrorno (Mean  = 3.00, S.E. = 0.21, \textit{p} < .05). There was no significant difference between \adequacyerroryes and \fluencyerrorno. 
There were no significant main effects for Spanish Proficiency or interactions between MT Error Type and Spanish Proficiency.

\paragraph{Recap \& Implications.} These results indicate that participants expressed lower trust in MT systems, as measured by their willingness to reuse the tool, after exposure to certain error types (\fluencyerrorno or \adequacyerroryes) but not others (\adequacyerrorno). 

This finding has practical implications. It suggests that controlled, fictional interactions with MT might influence future tool use, even through a brief session. Such settings make it possible to control for the impact of errors, facilitating risk awareness without waiting for natural errors. 
This can form the basis for future MT literacy interventions, motivating MT and NLP technical work to support their development. For example, semi-automatically producing stimuli, such as scoring or generating appropriate MT input/output pairs for a given context, decision, and error type, would be beneficial. More broadly, this highlights the need for system developers to understand technology's failure modes, whether for AI systems generally \citep{ehsan-etal-2022-seamful} or MT specifically \citep{robertson-etal-2023-angler}, as a complement to improving benchmark performance.
 
\subsection{RQ4: Individual Participants' Strategy Used to Evaluate MT Quality} 
\label{sec:results:rq4}

Last, we analyzed the main strategy reported by participants segmented by their Spanish Proficiency. We built a multinomial logistic regression model with the strategy categories as the dependent variable and MT Error Type and Spanish proficiency as predictors, while controlling for covariates as in previous models. In this model, No Strategy was set as the reference category for comparison.

Results reveal several significant predictors of strategy choice. Participants in \fluencyerrorno condition were significantly less likely to report using the \compstrategy (Coefficient = -2.30, SE = 0.26, \textit{p} < .001) and \profstrategy (Coefficient = -0.95, SE = 0.34, \textit{p} < .01).  Participants in \adequacyerrorno condition were also less likely to report the use of \compstrategy (Coefficient = -1.95, SE = 0.23, \textit{p} < .001) and \profstrategy (Coefficient = -1.39, SE = 0.41, \textit{p} = .001). Participants in the \adequacyerroryes condition were more likely to report \compstrategy (Coefficient = 1.78, SE = 0.31, \textit{p} < .001) but less likely to report \profstrategy(Coefficient = -1.80, SE = 0.33, \textit{p} < .001).
Spanish proficiency was found to have a significant effect on strategy choice. Participants with higher Spanish proficiency were less likely to report using \compstrategy (Coefficient = -5.34, SE = 0.44, \textit{p} < .001), but more likely to report using \profstrategy (Coefficient = 9.97, SE = 0.25, \textit{p} < .001).


\paragraph{Recap \& Implications.} These results indicate that different error conditions may prompt different strategies for evaluating translation quality. As expected, participants with higher Spanish proficiency are able to directly assess MT quality. More interestingly, participants attempt to compare the MT and the original more when they experience the negative impact of adequacy errors by losing points. However, the decision and confidence measures (Section~\ref{sec:results:rq12}) show that they are not successful when they are not proficient in Spanish. This calls for more work to design techniques that helps people make such comparisons. Providing cognitive forcing functions to encourage users to engage in a strategic evaluation of outputs has shown promise for other AI-decision making tasks \citep{buccinca2021trust}, and it remains to be seen how to design MT specific solutions. A promising strategy is to leverage question-answering to surface inconsistencies between the source and its translation \citep{ki-etal-2025-askqe, fernandes2025do}, which has been shown to help users decide whether a translation is reliable enough to share safely \citep{ki2025isharetranslationevaluating}. However, it remains unclear whether such feedback also supports users in making more accurate inferences when answering content-specific questions, as examined in this study.


\subsection{Post-Task Qualitative Feedback}
\label{sec:results:qualitative}

Post-task debrief sessions with participants indicated that the study prompted them to want to know more about MT. Notes taken by research assistants frequently mentioned the following debrief topics: participants' personal experience using MT for work and travel, understanding of the technical mechanisms of MT and why MT makes errors, participants' evaluation strategy used in the task, explanation of the translation errors in our stimuli, and general questions about our research design. Although we did not directly measure the educational benefits of this research participation experience, these discussion topics indicated a general interest of the public in acquiring information on MT systems following participation in the study. 

This feedback underscores a key research gap: supporting users in developing appropriate mental models of MT tools. While prior work has largely focused on feedback about the quality of individual outputs, less attention has been given to helping users understand the broader capabilities and limitations of these systems. Our findings point to the potential of simulations to promote MT literacy outside classroom settings. Although museum visitors who opt into such studies may be more motivated to learn than randomly selected MT users, this approach remains promising for professionals who rely on MT for communication with little or no formal training \citep{nunesvieira-2024-uses,mehandru-etal-2022-reliablea}. A more ambitious long-term goal is to design generic translation apps that explicitly foster MT literacy for all users.



%% file: latex/05_conclusion.tex
\section{Conclusion}

In summary, we presented a human-study investigating how fluency and adequacy errors impact bilingual and non-bilingual users’ reliance on MT during casual use. Our findings confirm that bilingual and non-bilingual users perceive and rely on potentially imperfect MT outputs differently. More surprisingly, they suggest that non-bilingual users over-rely on MT  not because they have high confidence in their correctness, but because they do not know what else to do: they lack strategies to evaluate outputs and reason about how to use them. However, experiencing the impact of errors in the study settings was sufficient to prompt users to reassess future reliance. 

These findings motivate several directions for future MT and NLP research, including the development of MT systems that support users in assessing and recovering from errors, the development of tools to support MT literacy training inspired by the task conducted here, as well as understanding how trust formation in casual settings common to everyday users \citep{vieira-etal-2022-machine} impact their behaviors in high-stakes use cases.

More broadly, we aim to underscore the value of complementing intrinsic evaluations of MT quality with studies of how people experience and respond to MT, thereby motivating future work at the intersection of NLP, HCI, and Translation Studies \citep{carpuat2025interdisciplinaryapproachhumancenteredmachine}. We also highlight the value of conducting human studies beyond the lab and crowdsourcing platforms to engage diverse segments of the public, using data collection as an opportunity for science communication \citep{vaughn-etal-2024-language} and promoting AI literacy.




%% file: latex/06_acknowledgements.tex
\section*{Acknowledgements}

We would like to thank the Language Science Station team and all the research assistants for hosting our research and facilitating data collection with museum visitors, including Ge (Stella) Huang, Sarah Nam, Kelly Brennan, Ashley Chau, Jamie Flynn, Miriam Franklin-Grinkorn, Lyosha Genzel, Tessa Goldlust, Tzipporah Harker, Isabel Harris, Maliel Henry, Mikoalis Bedrosian Kajen, Melanie Kurtz, Eamon Maloney, Kennedy Nwosu, Jessica Orozco Contreras, Cristelle Torbeso, Stacey Torbeso, and Nichole Tramel.
We are also grateful to the Planet Word leadership and staff for opening their doors to our study, and to the museum visitors who generously shared their time and insights with us.
This work was supported in part by NSF Language Science Station Grant 2116959 and by the Institute for Trustworthy AI in Law and Society (TRAILS), which is supported by the National Science Foundation under Award No. 2229885. The views and conclusions contained herein are those of the authors and should not be interpreted as necessarily representing the official policies, either expressed or implied, of NSF or the U.S. Government. The U.S. Government is authorized to reproduce and distribute reprints for governmental purposes notwithstanding any copyright annotation therein.

%% file: figs/appendix_fulltask.tex
\begin{figure*}[h]
    \centering
    \includegraphics[width=0.45\textwidth]{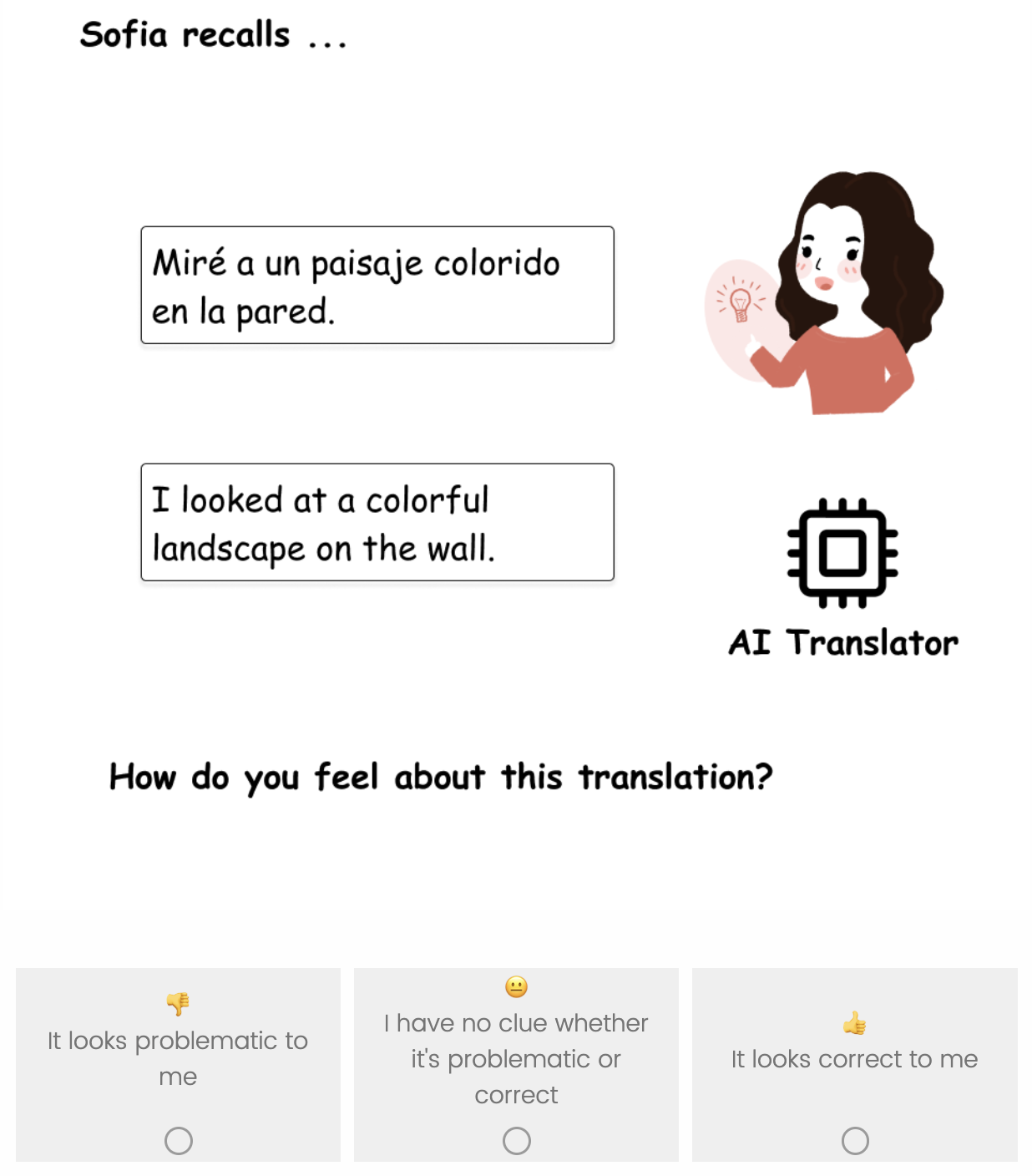}
    \includegraphics[width=0.45\textwidth]{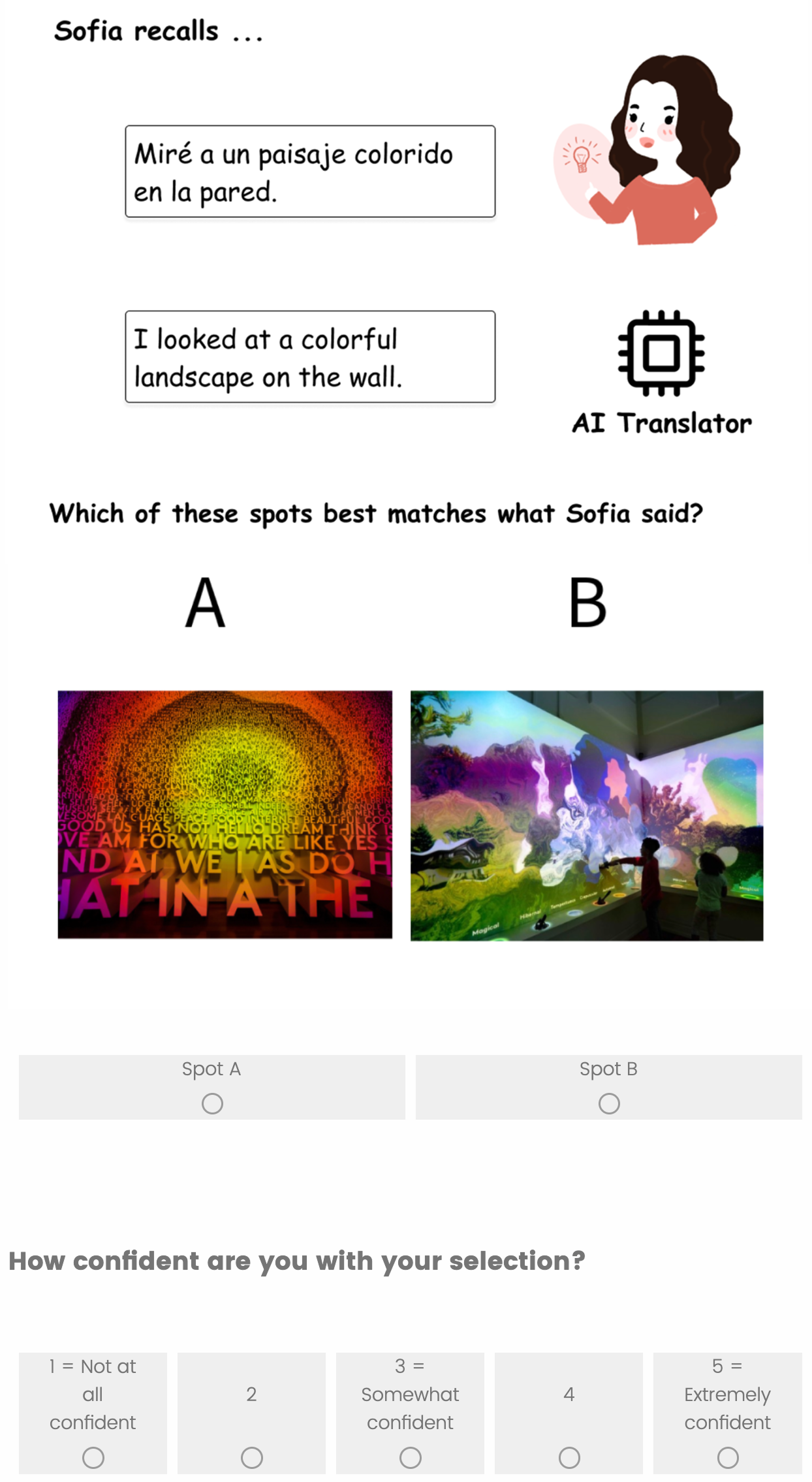}
    \caption{Screenshot of the full task for an example stimulus. \textbf{Left:} Participants rate their perception of the translation quality. \textbf{Right:} Next, participants select one out of two images that best match the location described in the Spanish message and rate their confidence in their selection. Image A shows colorful text radiating in a circular shape. Image B shows two children looking at a colorful, abstract digital display on the wall.}
    \label{fig:appendix_fulltask}
\end{figure*}

%% file: figs/appendix_fullstimuli.tex
\captionof{table}{All 16 stimuli used in our study. 
\textbf{Error Type:} Type of MT error present in the English MT; 
\textbf{Spanish Source:} Spanish source sentence; 
\textbf{English MT:} MT of the corresponding Spanish source; 
\textbf{Correct Image:} image describing the Spanish source; 
\textbf{Incorrect Image:} image that is plausible but does not describe the Spanish source.}
\label{tab:app_all_stimuli}

\begin{longtable*}{p{0.22\linewidth} p{0.22\linewidth} p{0.23\linewidth} p{0.23\linewidth}}

\toprule
\textbf{Spanish Source} & \textbf{English MT} & \textbf{Correct Image} & \textbf{Incorrect Image} \\
\midrule
\endfirsthead

\toprule
 \textbf{Spanish Source} & \textbf{English MT} & \textbf{Correct Image} & \textbf{Incorrect Image} \\
\midrule
\endhead

\midrule
\multicolumn{4}{r}{Continued on next page}\\
\endfoot

\bottomrule
\endlastfoot


\rowcolor{gray!15}
\multicolumn{4}{l}{\textbf{\correct}}\\
Miré a paisaje colorido en la pared. 
& I looked at a colorful landscape on the wall.
& \parbox[p]{0.22\textwidth}{
    \includegraphics[alt={Image displaying two children looking at a colorful, abstract digital display on the wall.}, width=0.22\textwidth]{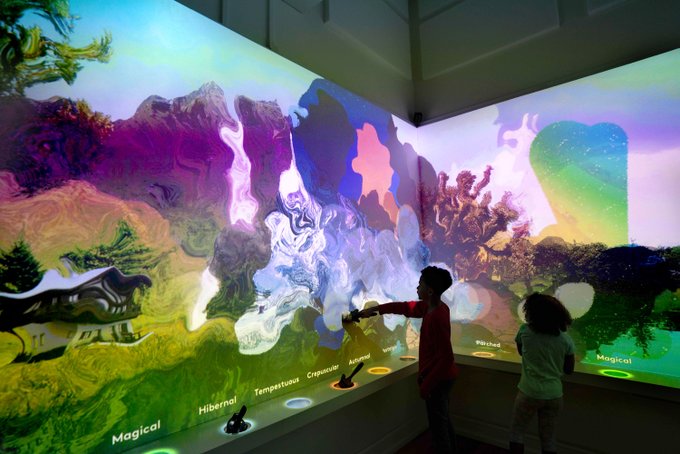} \\ 
    \footnotesize{(\textit{Two children looking at a colorful, abstract digital display on the wall.})}
    }
& \parbox[p]{0.22\textwidth}{
    \includegraphics[alt={Image displaying colorful text radiating in a circular shape.}, width=0.22\textwidth]{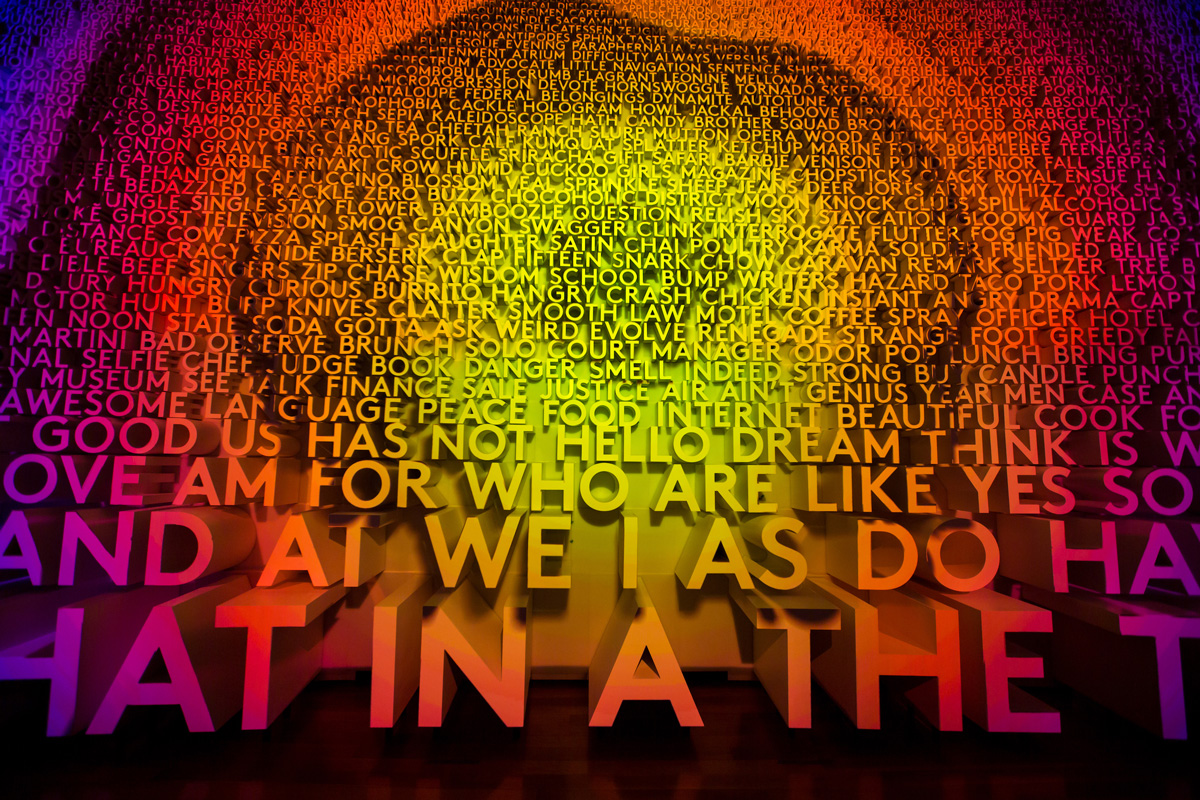} \\ 
    \footnotesize{(\textit{Colorful text radiating in a circular shape.})}
    } \\
\midrule

\rowcolor{gray!15}
\multicolumn{4}{l}{\textbf{\correct}}\\
 Me paré al lado de una cabina roja brillante.
& I stood next to a bright red booth.
& \parbox[p]{0.22\textwidth}{
    \includegraphics[alt={Image displaying red British-style telephone booth placed indoors on a polished wooden floor.}, width=0.22\textwidth]{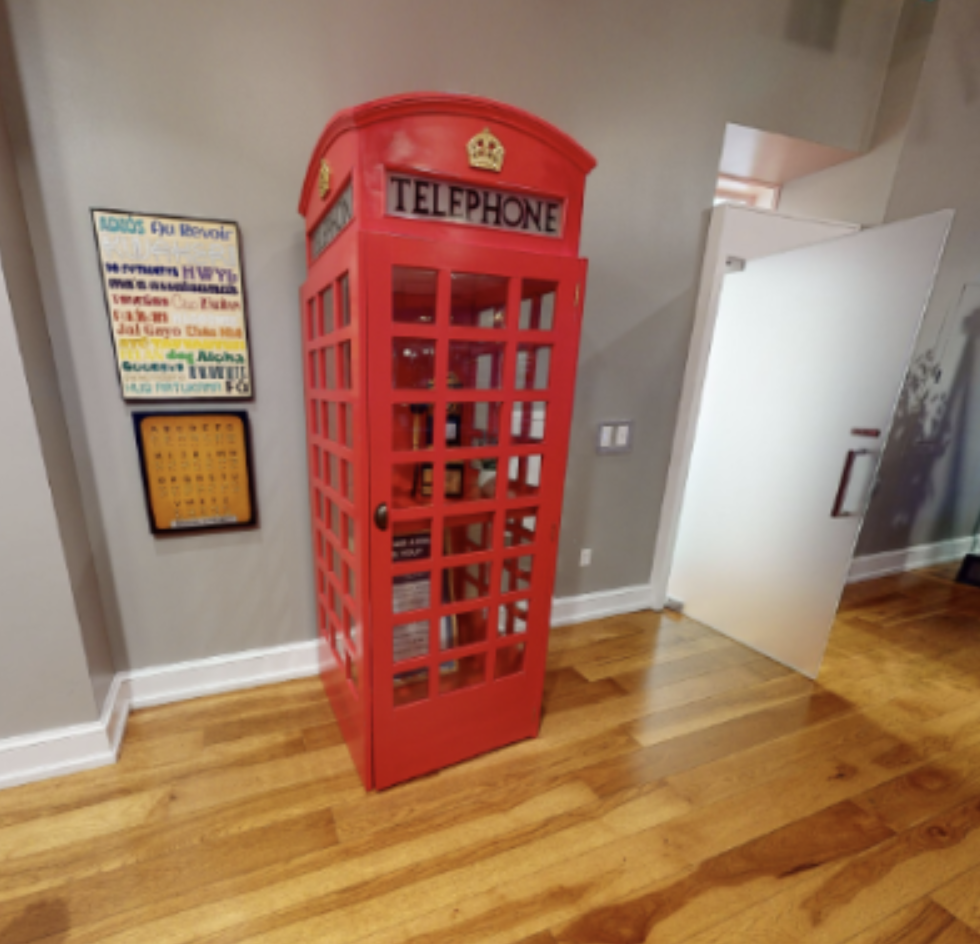} \\ 
    \footnotesize{(\textit{Red British-style telephone booth placed indoors on a polished wooden floor.})}
    }
    & 
    \parbox[p]{0.22\textwidth}{
        \includegraphics[alt={Image displaying photo booth with an arched entrance, red curtains, and a small stool inside.}, width=0.22\textwidth]{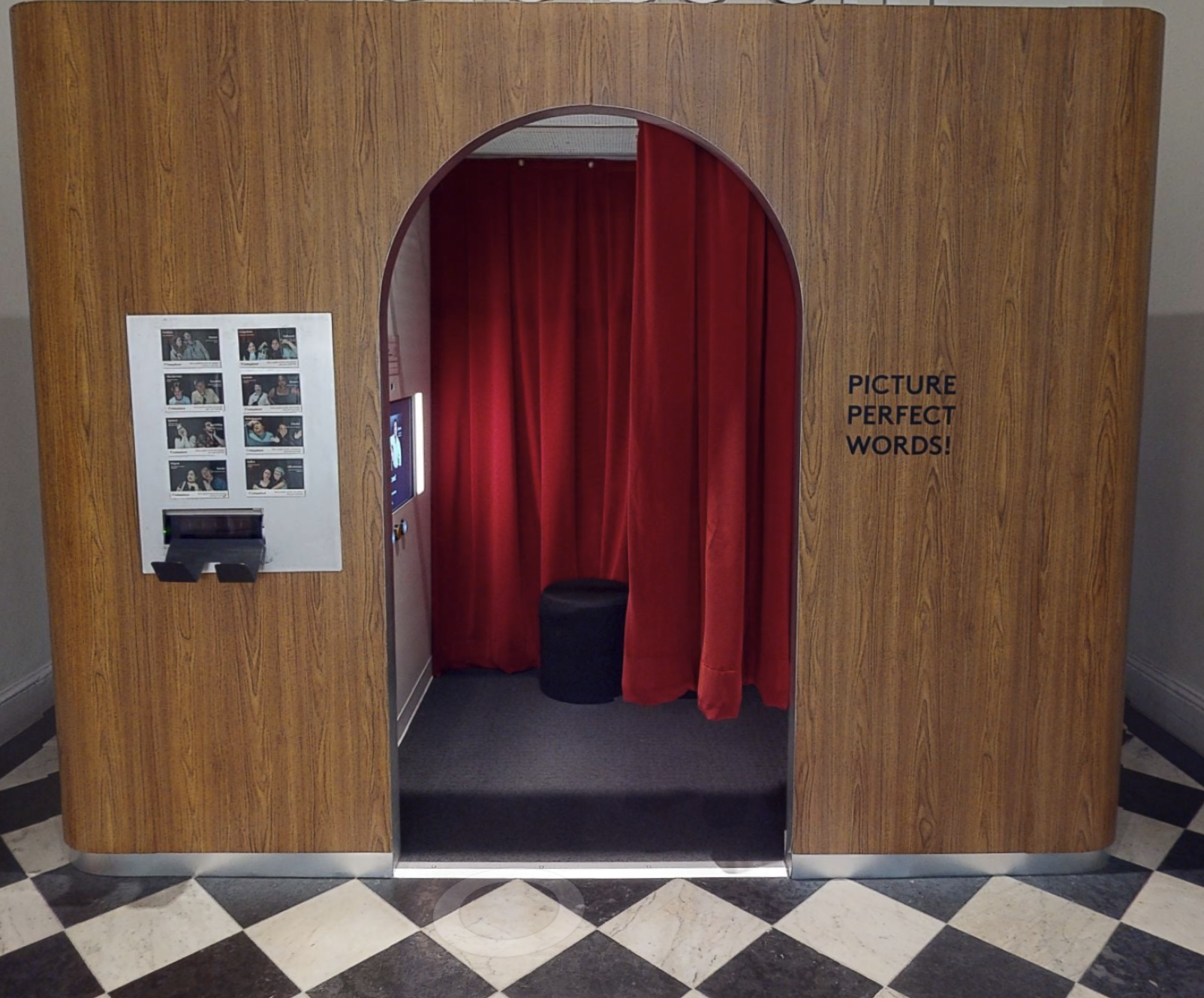} \\ 
        \footnotesize{(\textit{Photo booth with an arched entrance, red curtains, and a small stool inside.})}
    } \\
\midrule

\rowcolor{gray!15}
\multicolumn{4}{l}{\textbf{\fluencyerrorno}}\\
 Miré un instrumento de cuerda en una vitrina.
& Look at a bristle instrument in a showcase.
& \parbox[p]{0.22\textwidth}{
    \includegraphics[alt={Image displaying instruments behind the glass windows of a music shop with a blue storefront labeled ``Zithers Music Shop''.}, width=0.22\textwidth]{figs/appendix_imgs/s2_awkward1_correct.png} \\ 
    \footnotesize{(\textit{Instruments behind the glass windows of a music shop with a blue storefront labeled ``Zithers Music Shop''.})}
    }
& \parbox[p]{0.22\textwidth}{
    \includegraphics[alt={Image displaying karaoke setup on a wall, with the title ``Unlock the Music'' and visuals of lyrics and a man playing guitar.}, width=0.22\textwidth]{figs/appendix_imgs/s2_awkward1_incorrect.png} \\ 
    \footnotesize{(\textit{Karaoke setup on a wall, with the title ``Unlock the Music'' and visuals of lyrics and a man playing guitar.})}
    } \\
\midrule

\rowcolor{gray!15}
\multicolumn{4}{l}{\textbf{\fluencyerrorno}}\\
 Toqué los parlantes blancos que colgaban del árbol.
& I touched the white speakers hanging from the aol.
& \parbox[p]{0.22\textwidth}{
    \includegraphics[alt={Image displaying two women touching the white speakers hanging from a tree.}, width=0.2\textwidth]{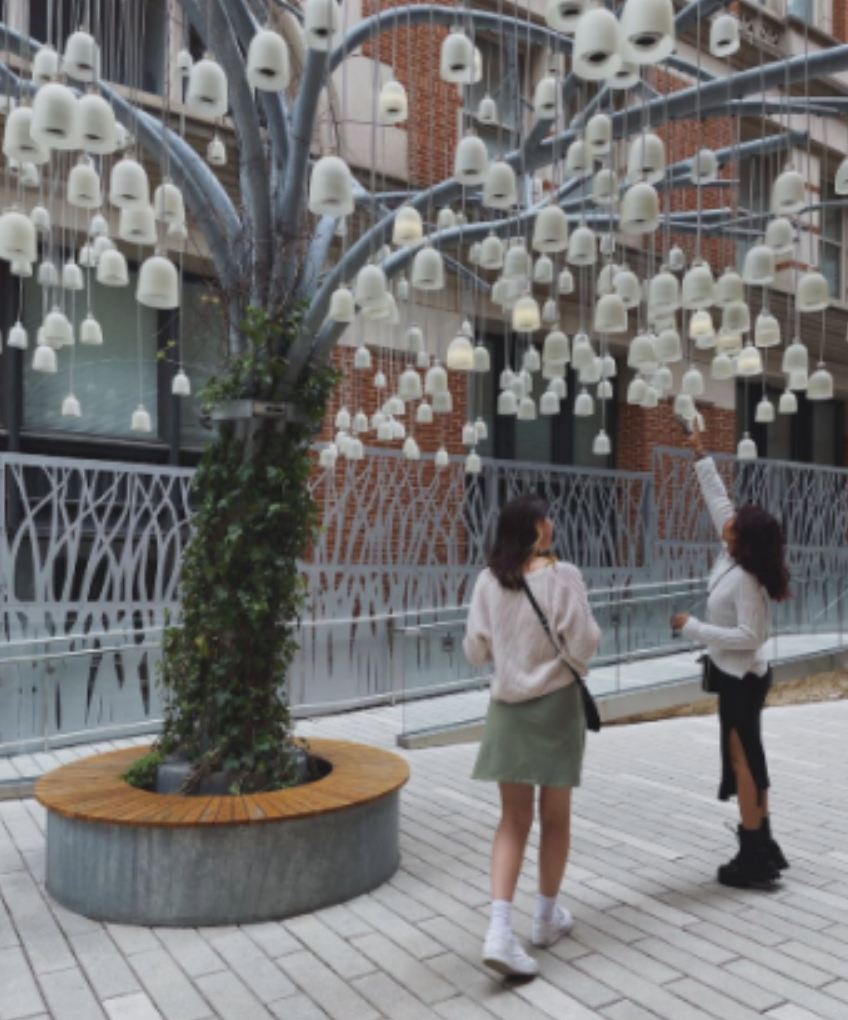} \\ 
    \footnotesize{(\textit{Two women touching the white speakers hanging from a tree.})}
    }
& \parbox[p]{0.22\textwidth}{
    \includegraphics[alt={Image displaying a purple wall displaying interactive screens with microphones handing on the top.}, width=0.22\textwidth]{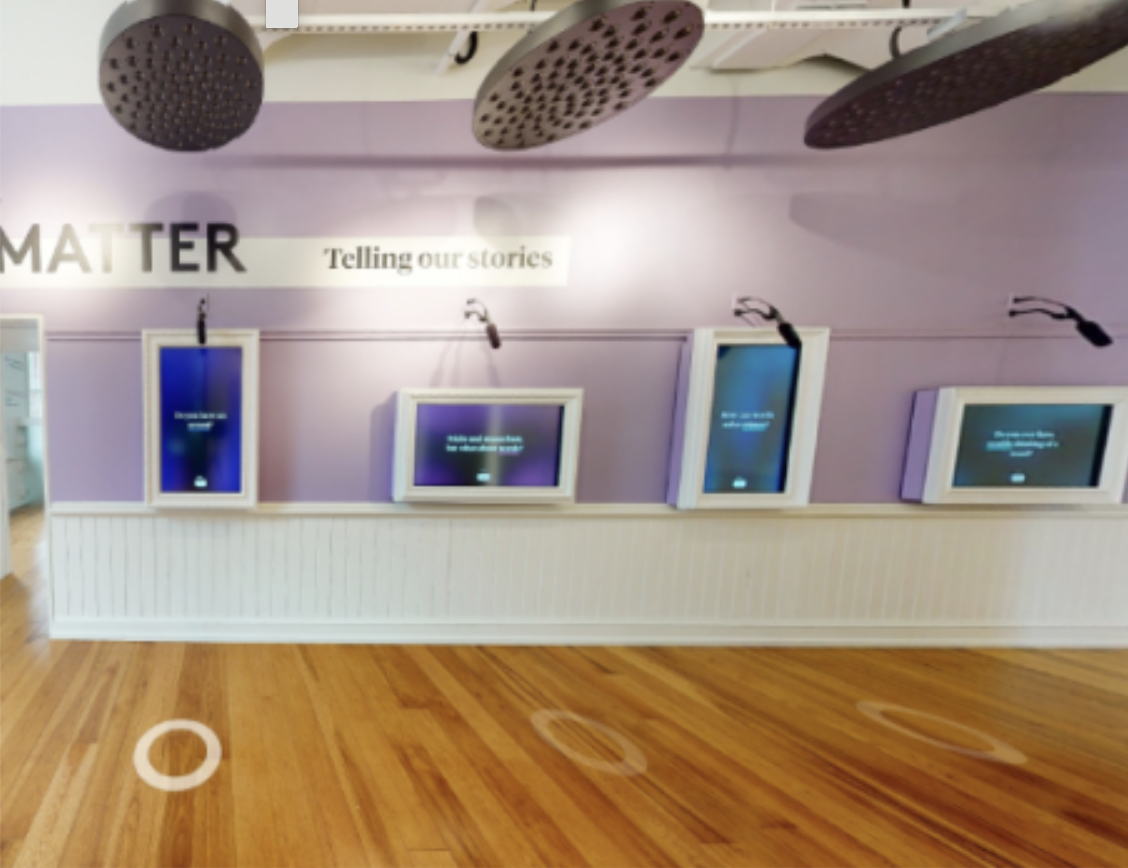} \\ 
    \footnotesize{(\textit{A purple wall displaying interactive screens with microphones handing on the top.})}
    } \\
\midrule

\rowcolor{gray!15}
\multicolumn{4}{l}{\textbf{\fluencyerrorno}}\\
 Miré a los libros para niños en la sala de la biblioteca.
& I looked at the children's bros in the library room.
& \parbox[p]{0.22\textwidth}{
    \includegraphics[alt={Image displaying a child wearing a blue mask engaging with an interactive display in a library.}, width=0.22\textwidth]{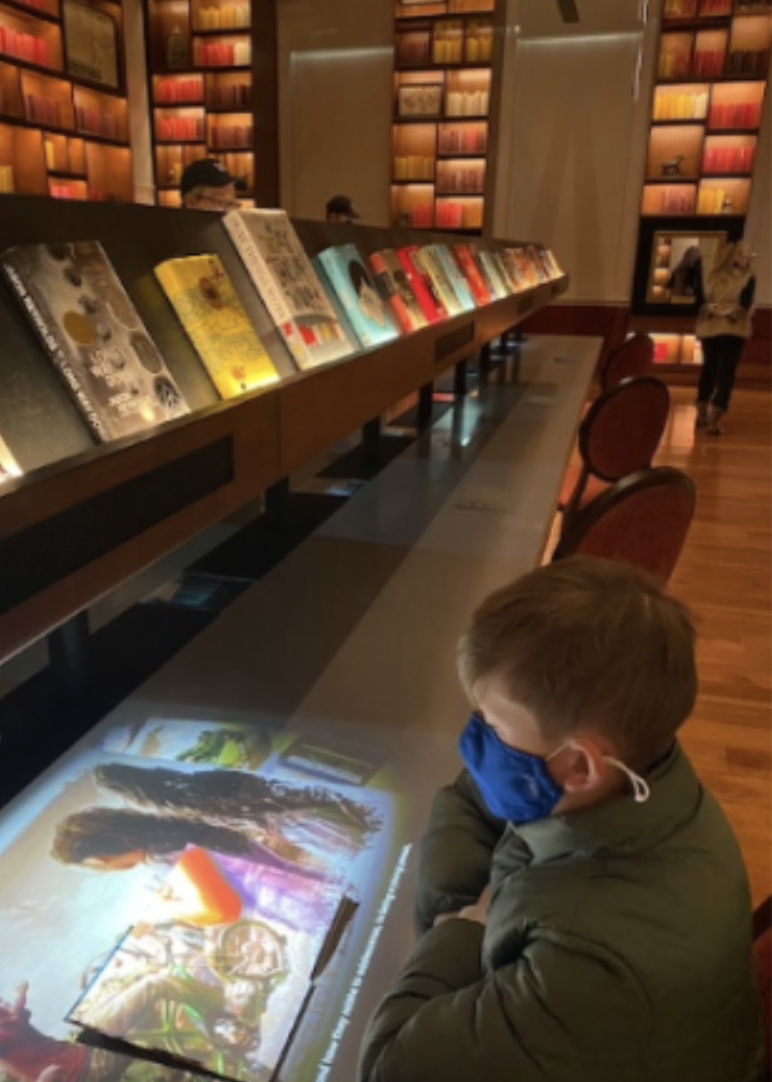} \\ 
    \footnotesize{(\textit{A child wearing a blue mask engaging with an interactive display in a library.})}
    }
& \parbox[p]{0.22\textwidth}{
    \includegraphics[alt={Image displaying a scrabble-themed merchandise, including books and decorative items in a gift shop.}, width=0.22\textwidth]{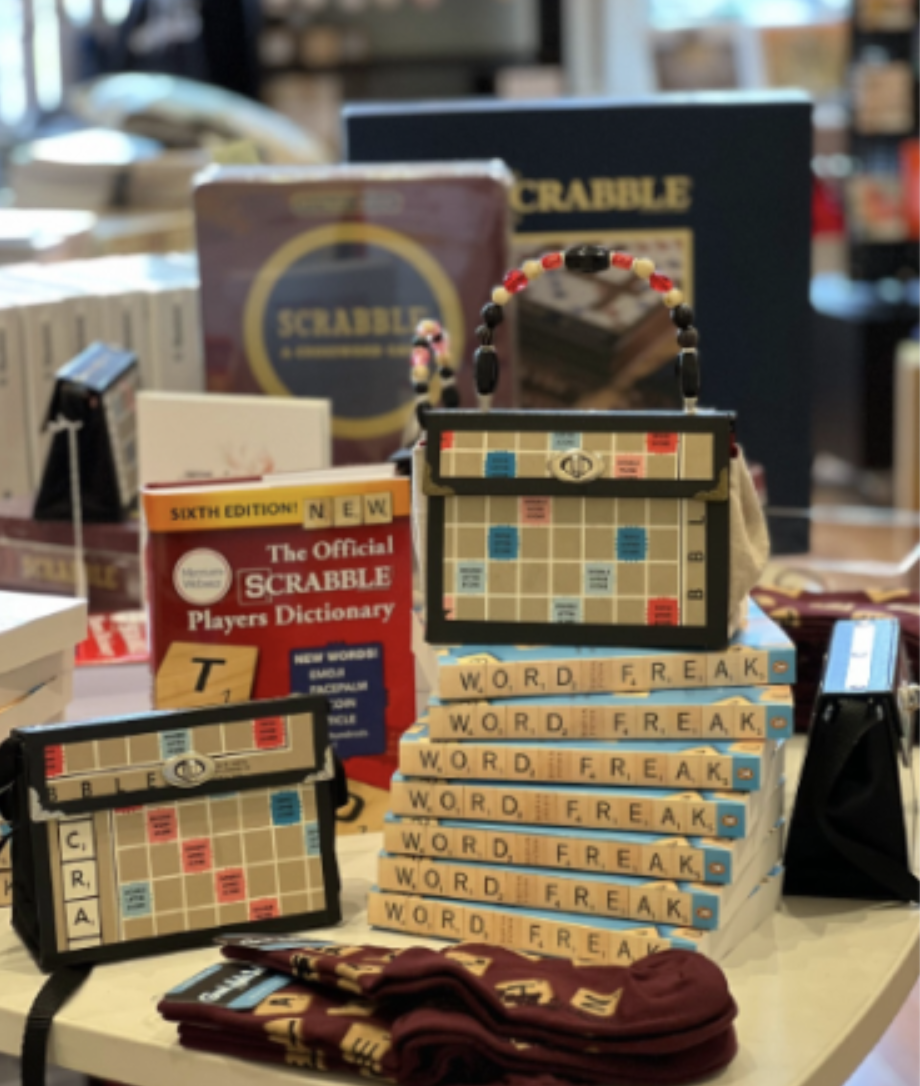} \\ 
    \footnotesize{(\textit{Scrabble-themed merchandise, including books and decorative items in a gift shop.})}
    } \\
\midrule

\rowcolor{gray!15}
\multicolumn{4}{l}{\textbf{\fluencyerrorno}}\\
 Escribí mis pensamientos sobre el libro en un nota adhesiva.
& I wrote my thoughts on the sticky book.
& \parbox[p]{0.22\textwidth}{
    \includegraphics[alt={Image displaying pink sticky notes on a white table along with a pen and a laptop.}, width=0.22\textwidth]{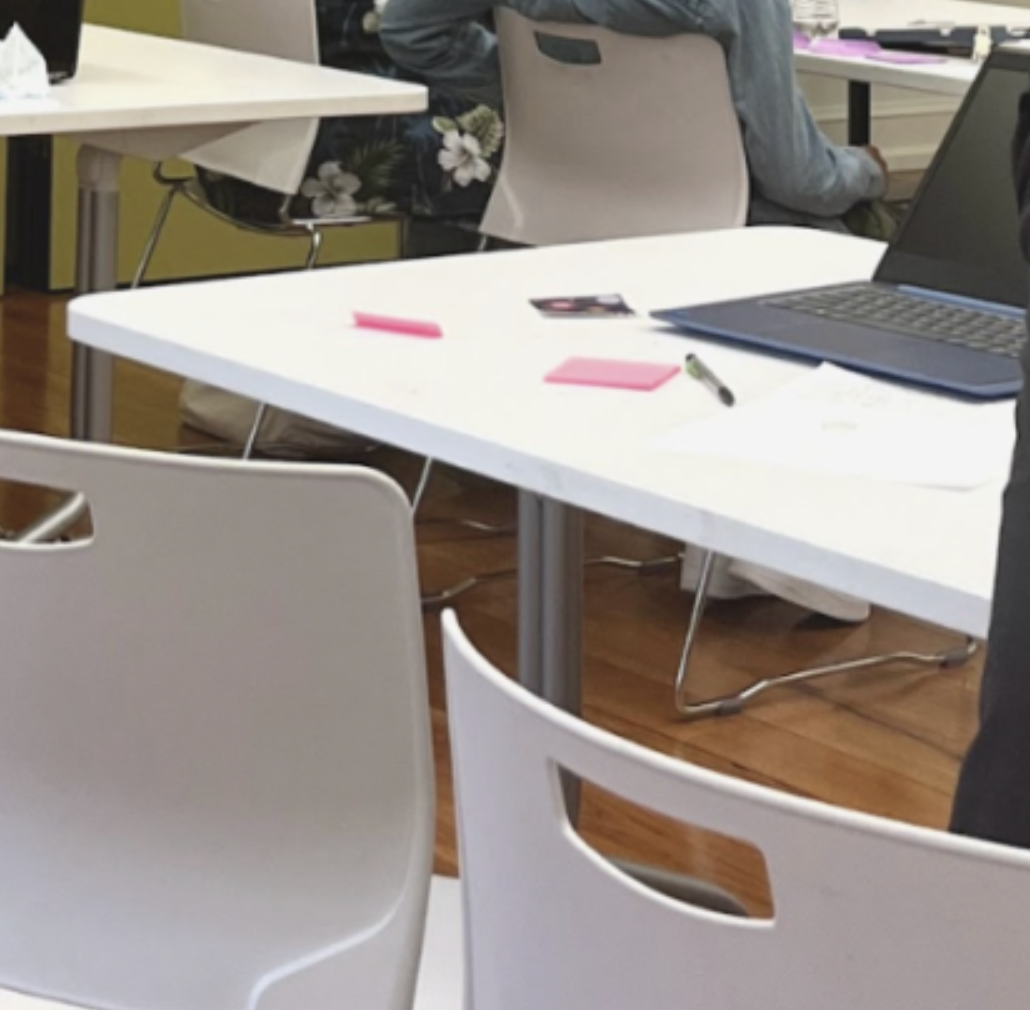} \\ 
    \footnotesize{(\textit{Pink sticky notes on a white table along with a pen and a laptop.})}
    }
& \parbox[p]{0.22\textwidth}{
    \includegraphics[alt={Image displaying an illuminated open book with depictions of two figures surrounded by flames.}, width=0.22\textwidth]{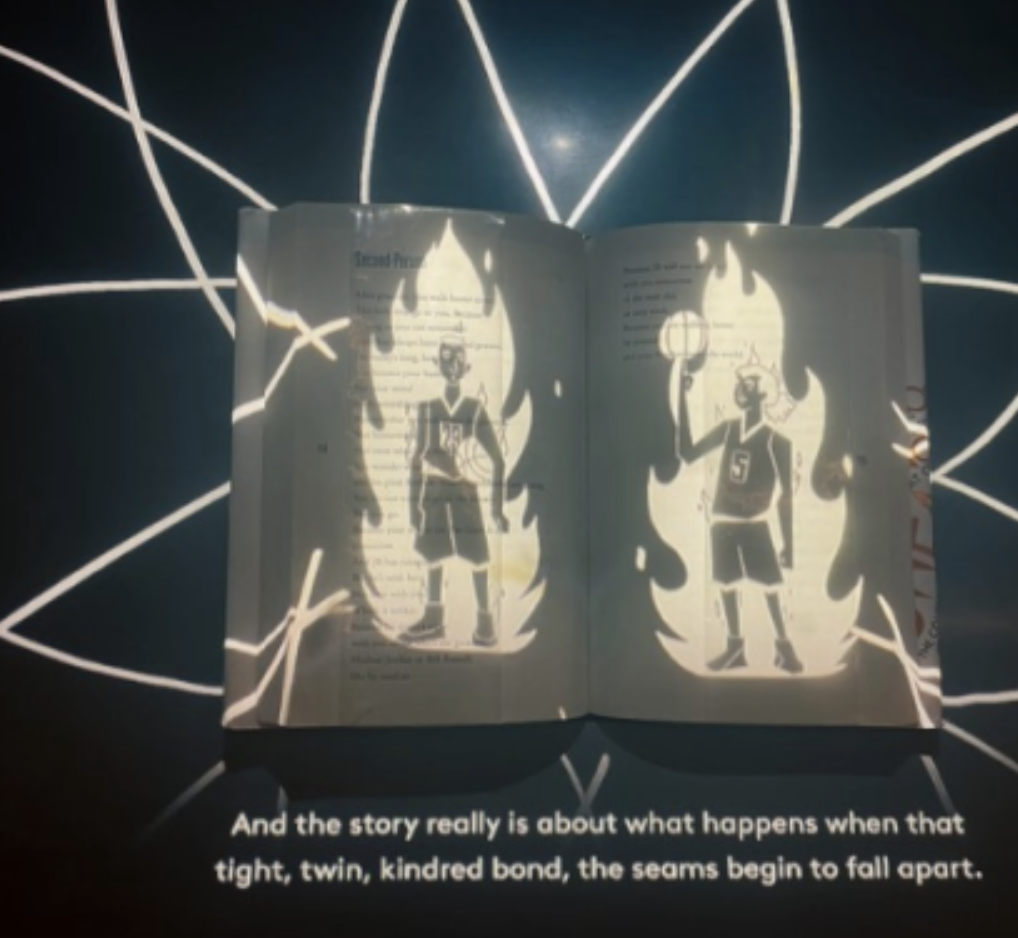} \\ 
    \footnotesize{(\textit{Illuminated open book with depictions of two figures surrounded by flames.})}
    } \\
\midrule

\rowcolor{gray!15}
\multicolumn{4}{l}{\textbf{\adequacyerrorno}}\\
Subí al tercer piso.
& Take it to the third floor.
& \parbox[p]{0.22\textwidth}{
    \includegraphics[alt={Image displaying stairs at the third floor with patterned tiles and decorative iron railings.}, width=0.22\textwidth]{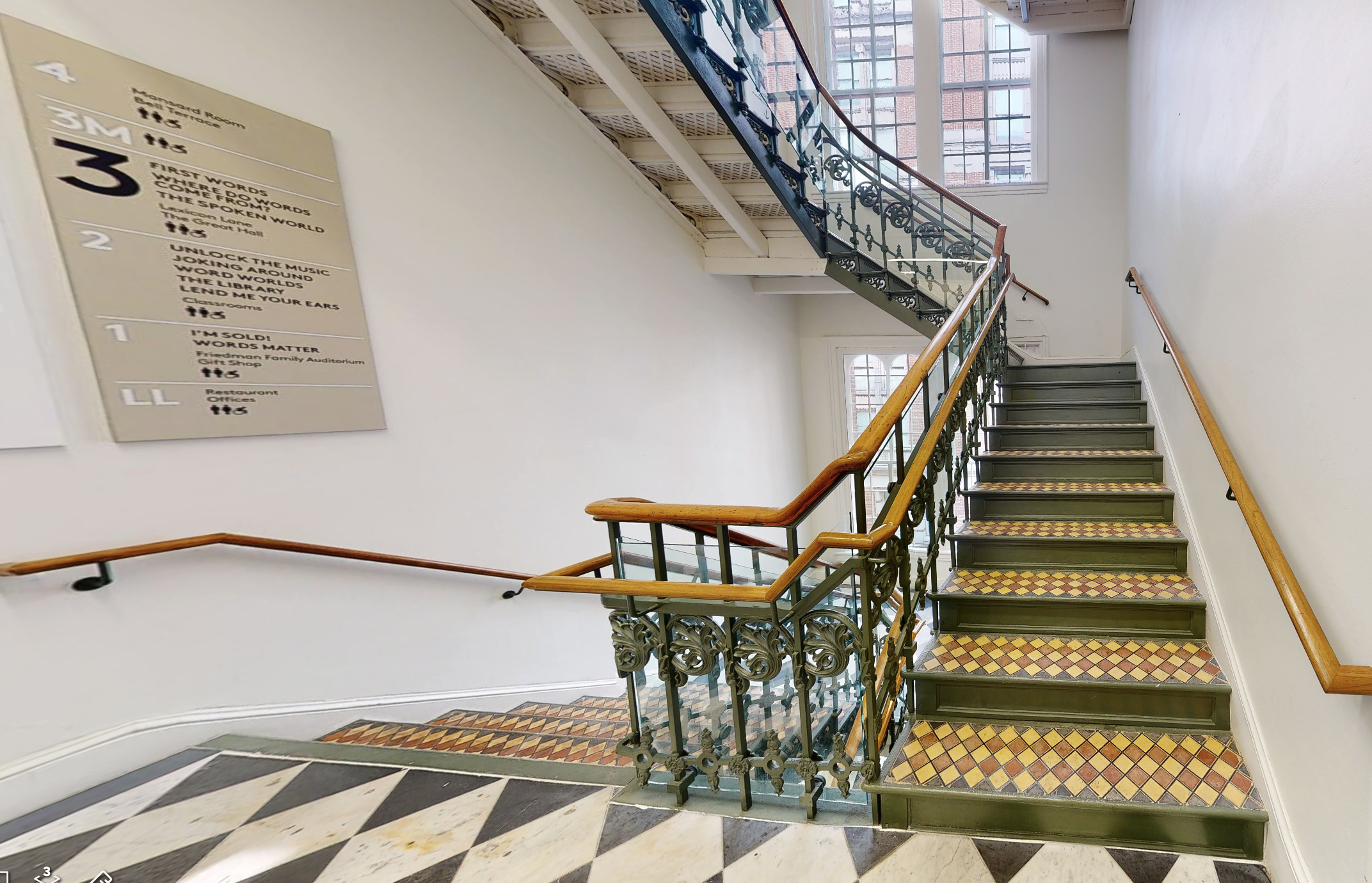} \\ 
    \footnotesize{(\textit{Stairs at the third floor with patterned tiles and decorative iron railings.})}
    }
& \parbox[p]{0.22\textwidth}{
    \includegraphics[alt={Image displaying directory sign at the second floor.}, width=0.22\textwidth]{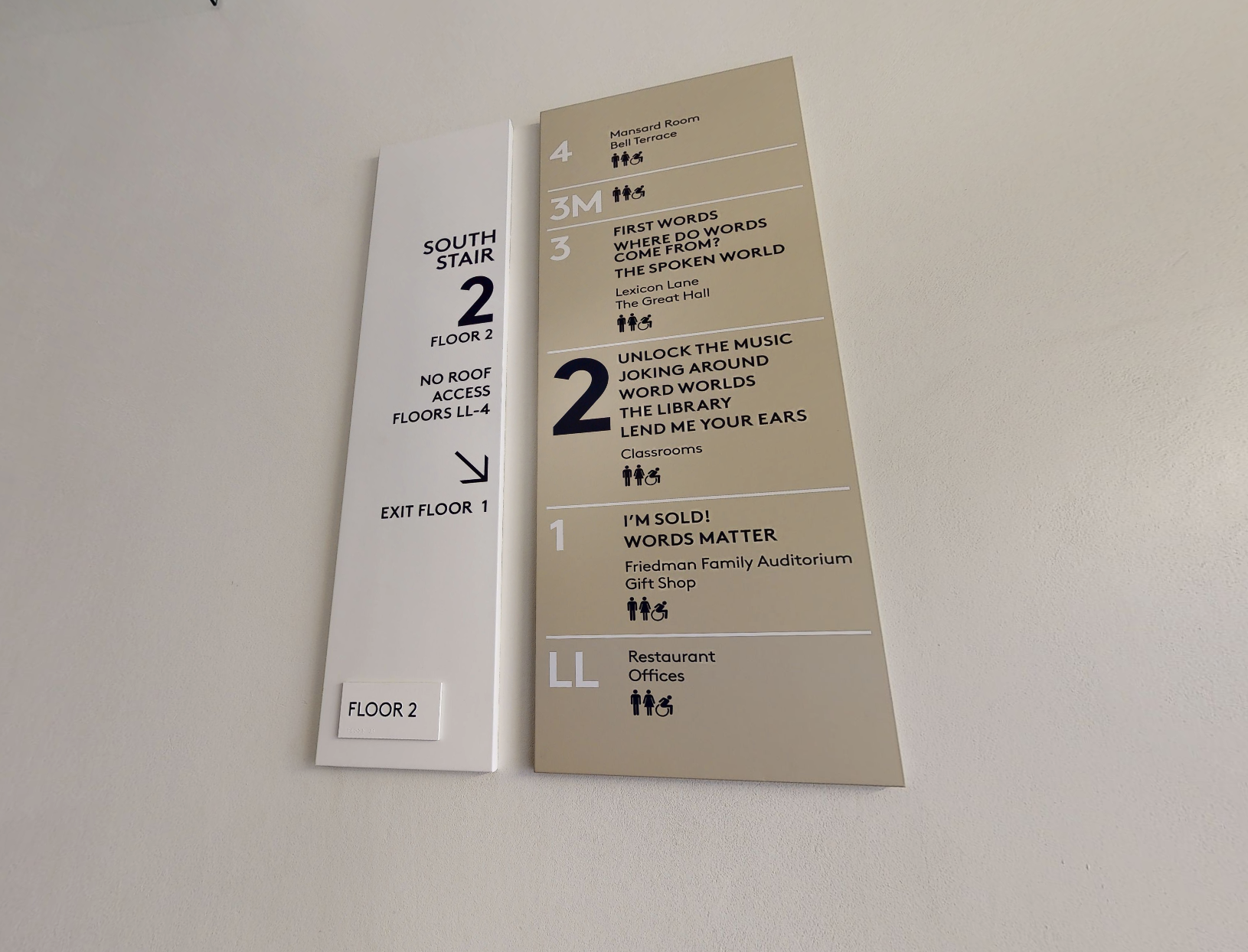} \\ 
    \footnotesize{(\textit{Directory sign at the second floor.})}
    } \\
\midrule

\rowcolor{gray!15}
\multicolumn{4}{l}{\textbf{\adequacyerrorno}}\\
Escuché los balbuceos de una pequeña bebé.
& I listened to the babbles of a little baby.
& \parbox[p]{0.22\textwidth}{
    \includegraphics[alt={Image displaying digital audio screen surrounded by framed photos of babies on a wall.}, width=0.22\textwidth]{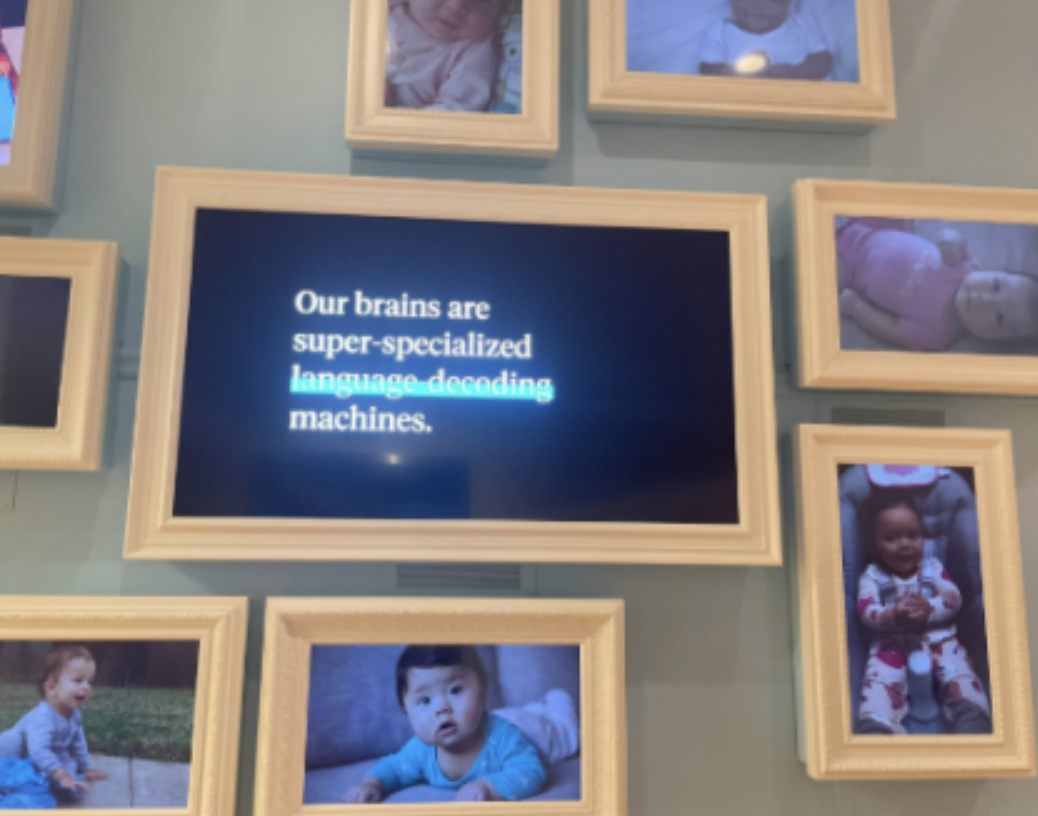} \\ 
    \footnotesize{(\textit{Digital audio screen surrounded by framed photos of babies on a wall.})}
    }
& \parbox[p]{0.22\textwidth}{
    \includegraphics[alt={Image displaying audience seated in a room watching a large screen with a picture of a smiling girl wearing glasses.}, width=0.22\textwidth]{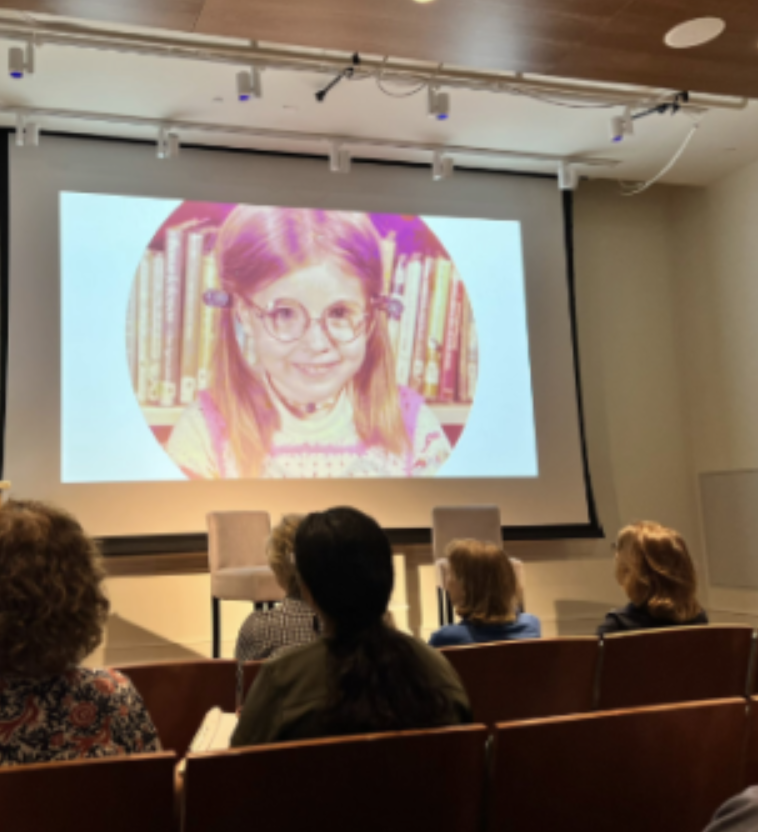} \\ 
    \footnotesize{(\textit{Audience seated in a room watching a large screen with a picture of a smiling girl wearing glasses.})}
    }
    \\
\midrule

\rowcolor{gray!15}
\multicolumn{4}{l}{\textbf{\adequacyerrorno}}\\
Habían varias pantallas chicas.
& There were several girls in glasses.
& \parbox[p]{0.22\textwidth}{
    \includegraphics[alt={Image displaying illuminated spherical globe with digital screens featuring women in glasses.}, width=0.22\textwidth]{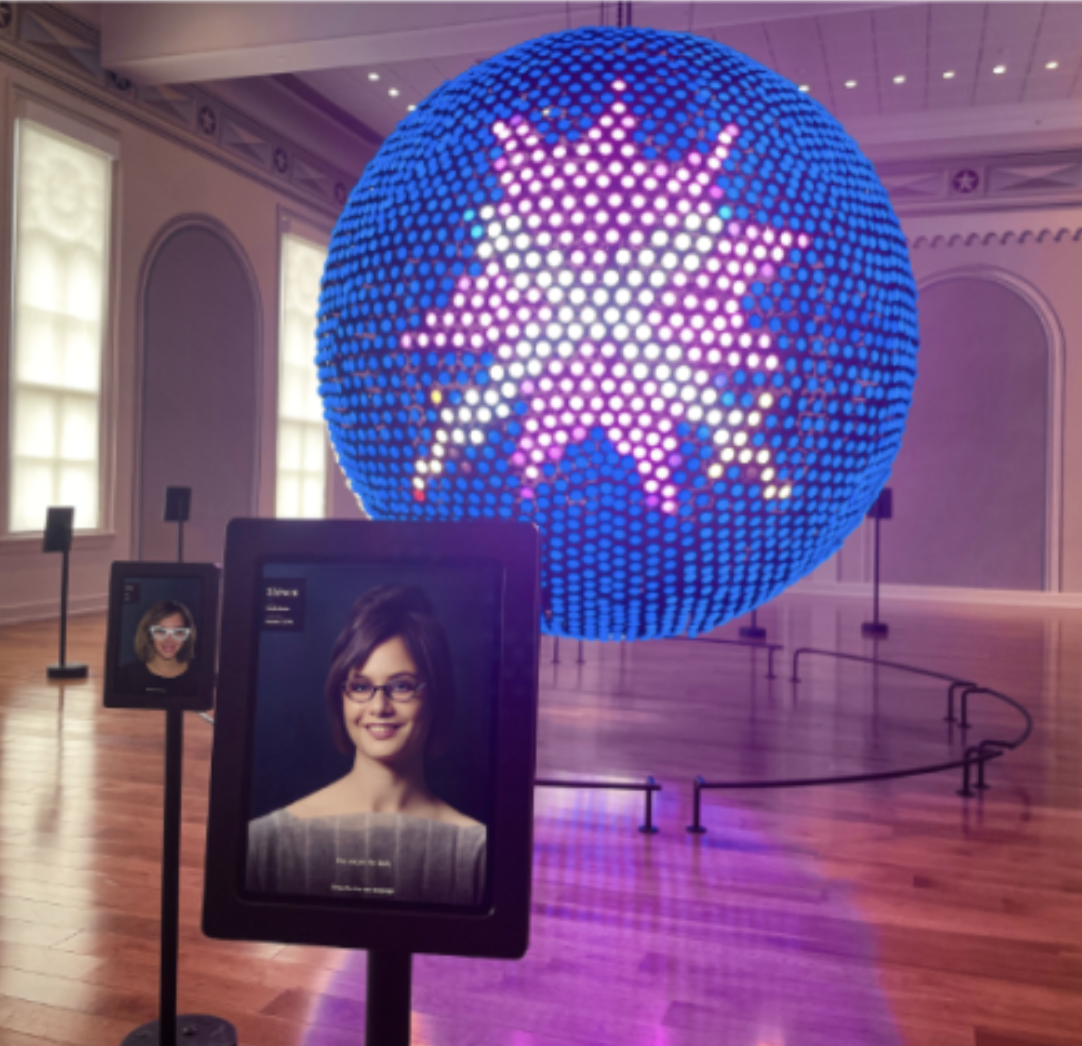} \\ 
    \footnotesize{(\textit{Illuminated spherical globe with digital screens featuring women in glasses.})}
    }
& \parbox[p]{0.22\textwidth}{
    \includegraphics[alt={Image displaying an exhibit with a projected video featuring a person discussing a topic ``Broken English''.}, width=0.22\textwidth]{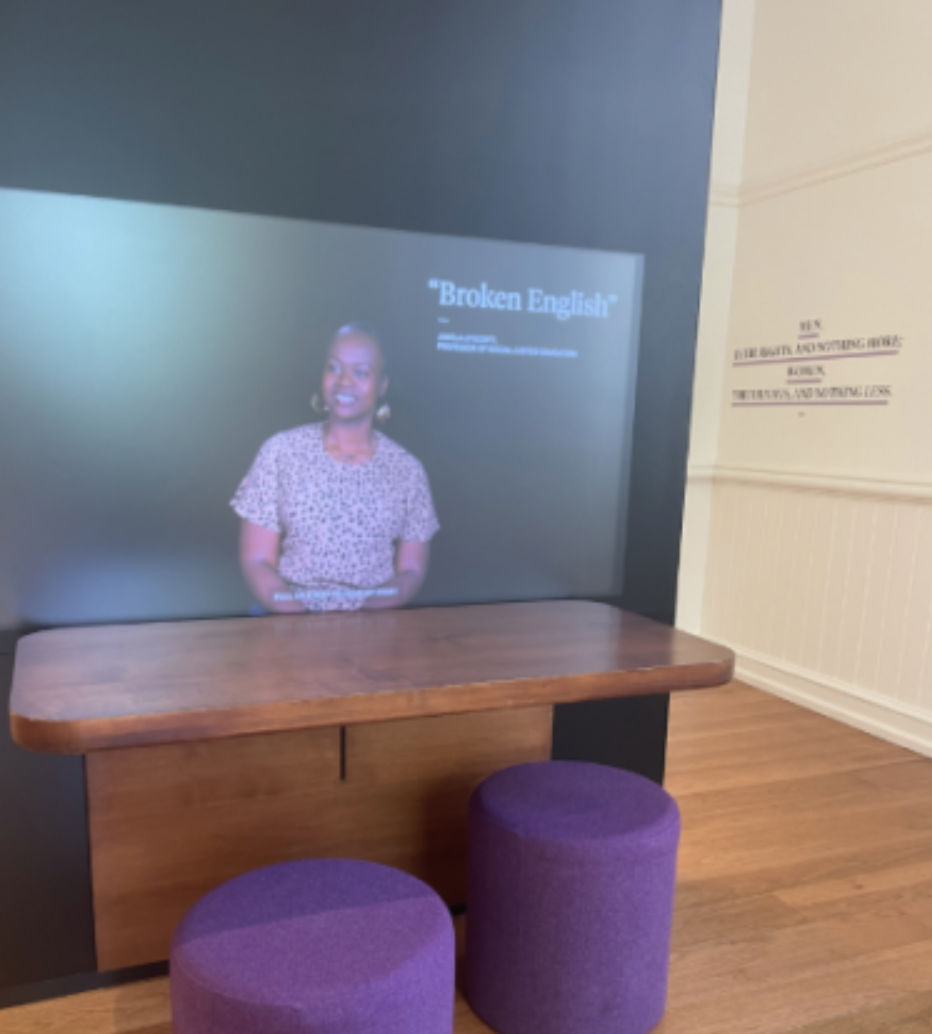} \\ 
    \footnotesize{(\textit{Exhibit with a projected video featuring a person discussing a topic ``Broken English''.})}
    } \\
\midrule

\rowcolor{gray!15}
\multicolumn{4}{l}{\textbf{\adequacyerrorno}}\\
 Pisé letras y señales de muchos idiomas.
& Write letters and signs of many languages.
& \parbox[p]{0.22\textwidth}{
    \includegraphics[alt={Image displaying floor near an elevator with characters and signs in many languages spread around.}, width=0.22\textwidth]{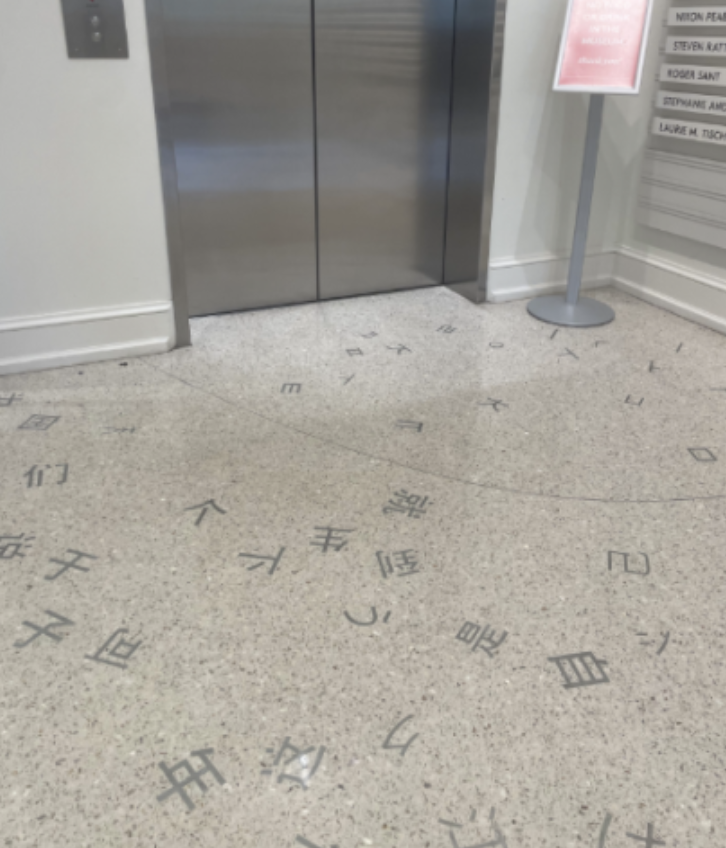} \\ 
    \footnotesize{(\textit{Floor near an elevator with characters and signs in many languages spread around.})}
    }
& \parbox[p]{0.22\textwidth}{
    \includegraphics[alt={Image displaying a girl looking at a newspaper on a flat digital screen.}, width=0.22\textwidth]{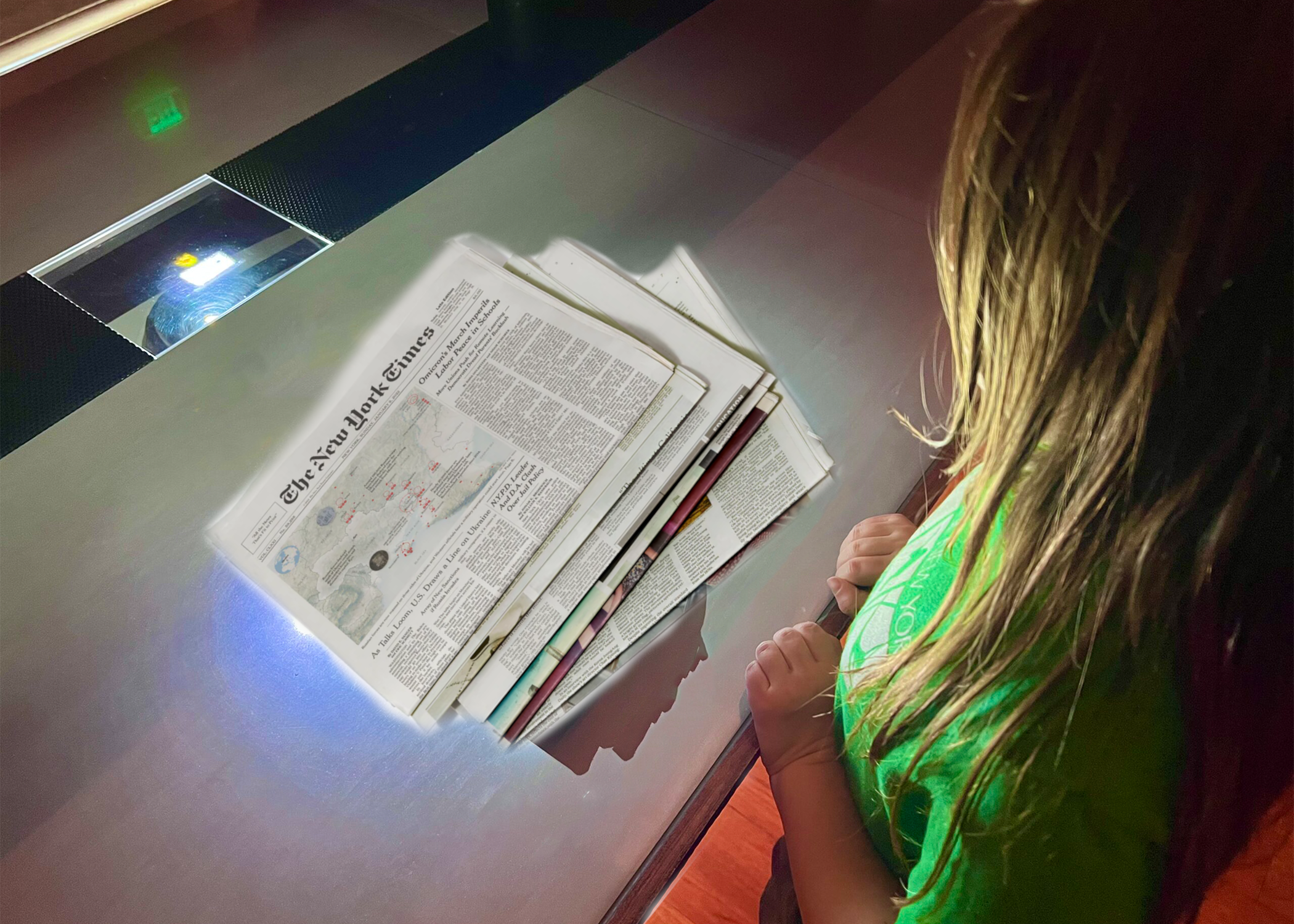} \\ 
    \footnotesize{(\textit{Girl looking at a newspaper on a flat digital screen.})}
    } \\
\midrule

\rowcolor{gray!15}
\multicolumn{4}{l}{\textbf{\adequacyerroryes}}\\
Pisé letras y señales de muchos idiomas.
& Write letters and signs of many languages.
& \parbox[p]{0.22\textwidth}{
    \includegraphics[alt={Image displaying floor near an elevator with characters and signs in many languages spread around.}, width=0.22\textwidth]{figs/appendix_imgs/s2_adeq_w1_correct.png} \\ 
    \footnotesize{(\textit{Floor near an elevator with characters and signs in many languages spread around.})}
    }
& \parbox[p]{0.22\textwidth}{
    \includegraphics[alt={Image displaying green sticky notes on a wall with the text ``One word I love from another language is ...''.}, width=0.22\textwidth]{figs/appendix_imgs/s2_adeq_w1_incorrect.png} \\ 
    \footnotesize{(\textit{Green sticky notes on a wall with the text ``One word I love from another language is ...''.})}
    } \\
\midrule

\rowcolor{gray!15}
\multicolumn{4}{l}{\textbf{\adequacyerroryes}}\\
 Vi una historia cobrar vida.
& I saw history come to life.
& \parbox[p]{0.22\textwidth}{
    \includegraphics[alt={Image displaying illuminated open book with silhouettes of animals, trees, and people.}, width=0.22\textwidth]{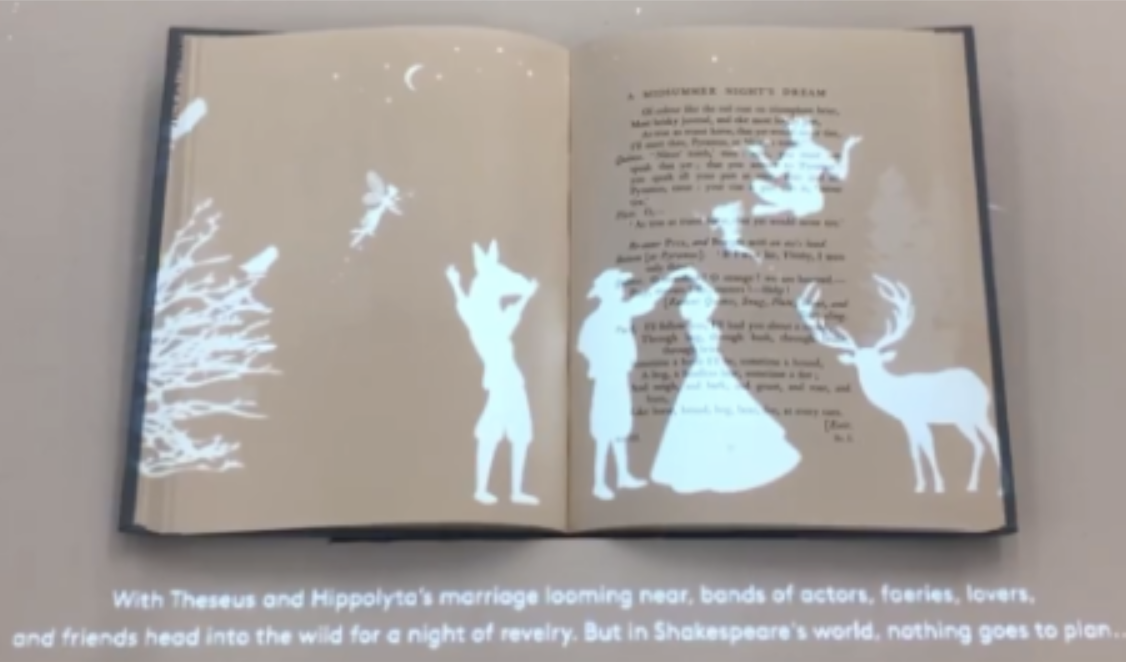} \\ 
    \footnotesize{(\textit{Illuminated open book with silhouettes of animals, trees, and people.})}
    }
& \parbox[p]{0.22\textwidth}{
    \includegraphics[alt={Image displaying framed screen displaying a historical speech.}, width=0.22\textwidth]{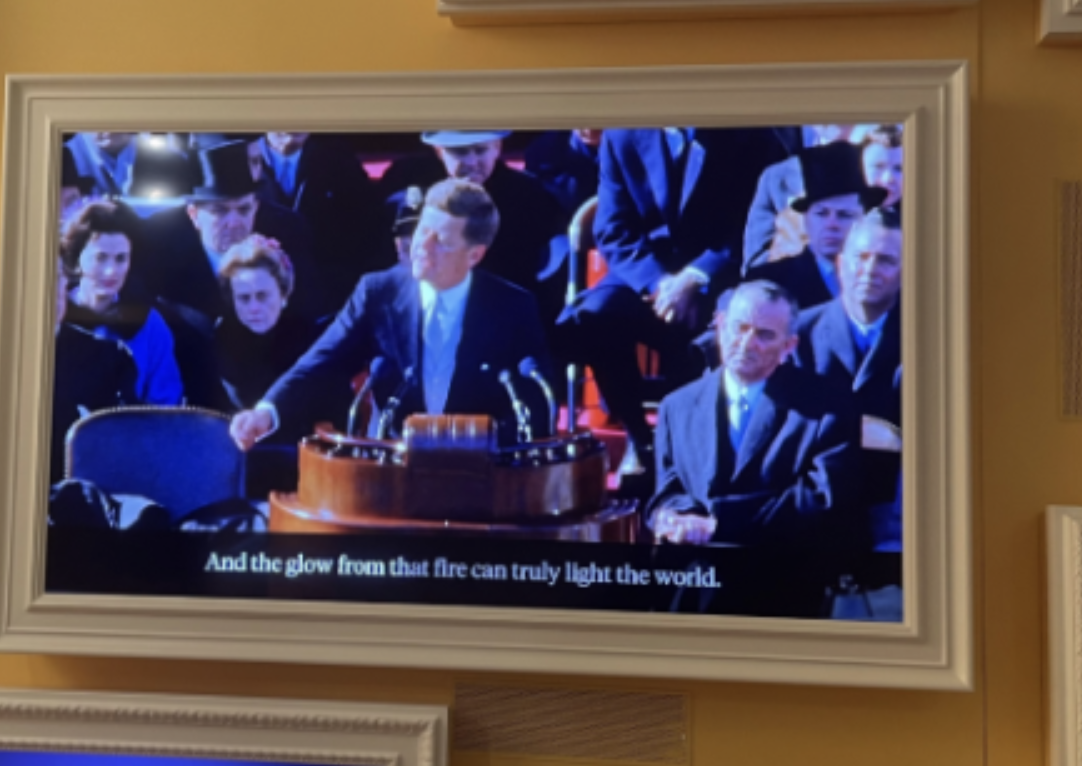} \\ 
    \footnotesize{(\textit{Framed screen displaying a historical speech.})}
    } \\
\midrule

\rowcolor{gray!15}
\multicolumn{4}{l}{\textbf{\adequacyerroryes}}\\
 Había varias pantallas chicas.
& There were several girls in glasses.
& \parbox[p]{0.22\textwidth}{
    \includegraphics[alt={Image displaying display with multiple framed screens against a light blue wall.}, width=0.22\textwidth]{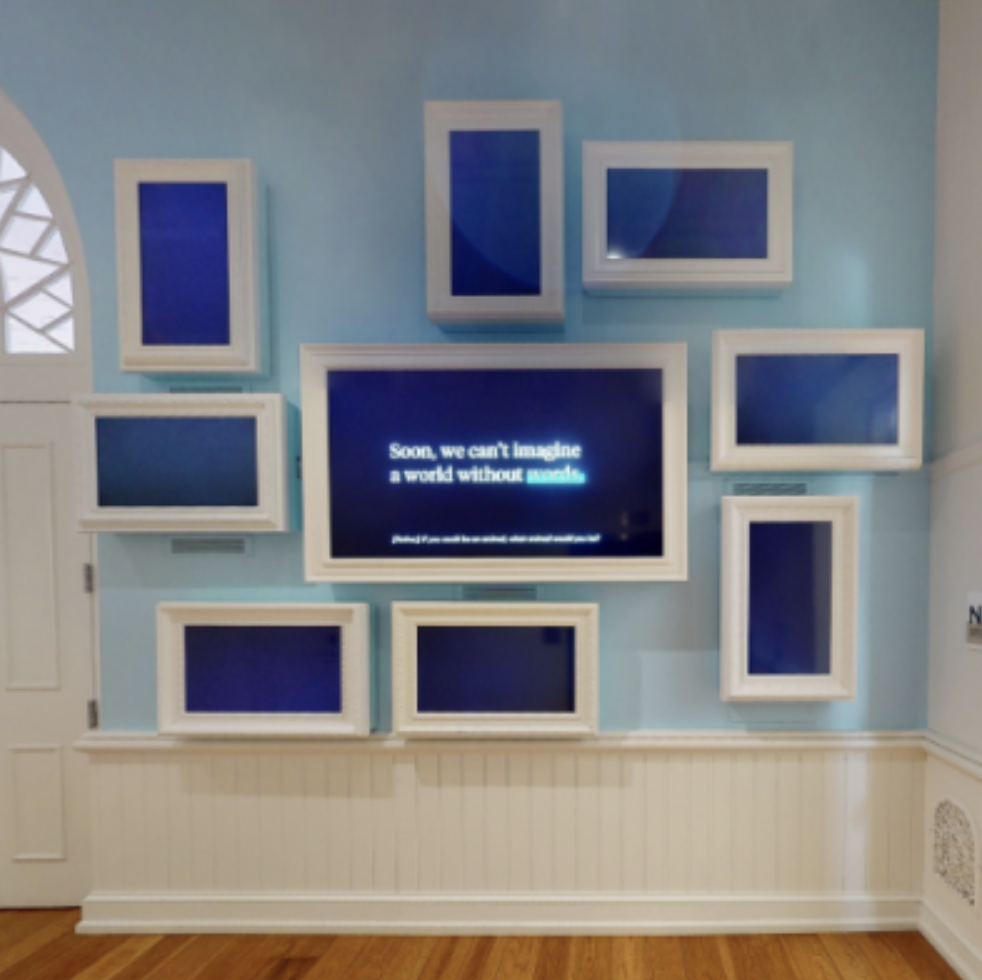} \\ 
    \footnotesize{(\textit{Display with multiple framed screens against a light blue wall.})}
    }
& \parbox[p]{0.22\textwidth}{
    \includegraphics[alt={Image displaying audience seated in a room watching a large screen with a picture of a smiling girl wearing glasses.}, width=0.22\textwidth]{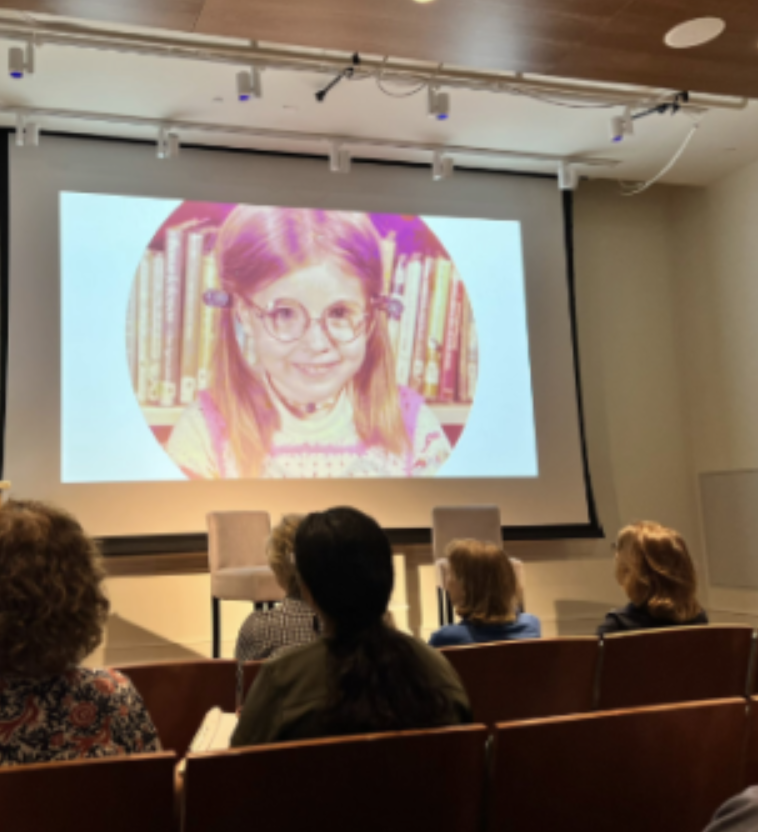} \\ 
    \footnotesize{(\textit{Audience seated in a room watching a large screen with a picture of a smiling girl wearing glasses.})}
    } \\
\midrule

\rowcolor{gray!15}
\multicolumn{4}{l}{\textbf{\adequacyerroryes}}\\
 Me senté en el asiento al frente del auditorio.
& I sat in the seat at the front of the classroom.
& \parbox[p]{0.22\textwidth}{
    \includegraphics[alt={Image displaying front view of the classroom-style seats with teal-cushioned chairs.}, width=0.22\textwidth]{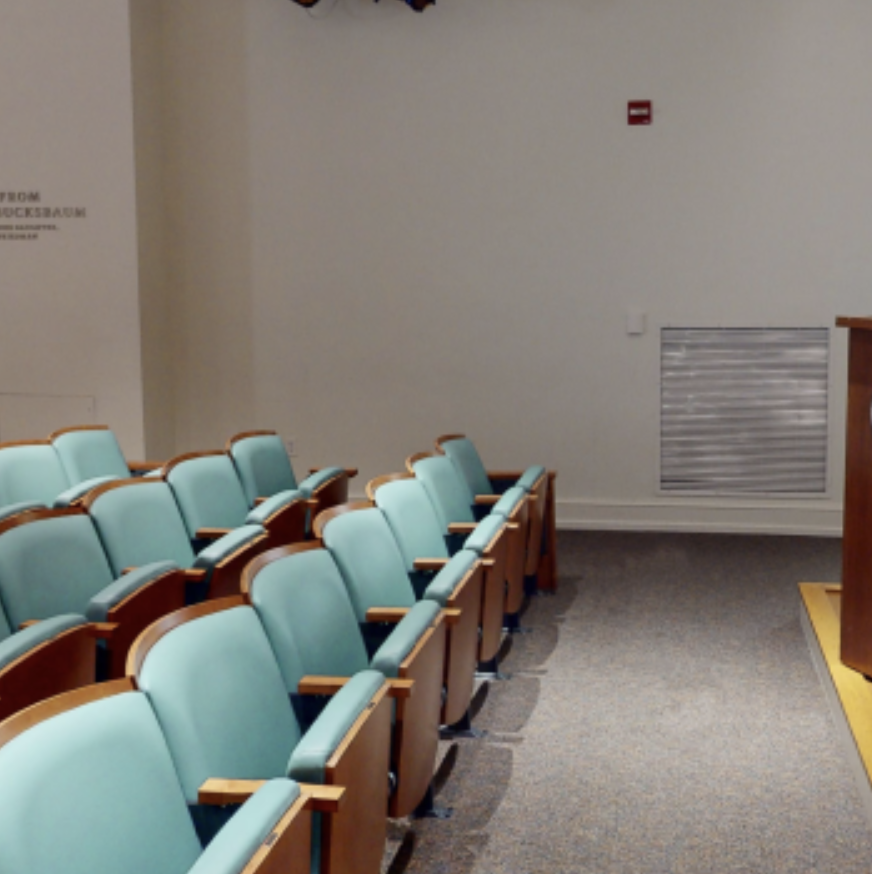} \\ 
    \footnotesize{(\textit{Front view of the classroom-style seats with teal-cushioned chairs.})}
    }
& \parbox[p]{0.22\textwidth}{
    \includegraphics[alt={Image displaying second row view of the classroom with white tables facing a wall with blackboards.}, width=0.22\textwidth]{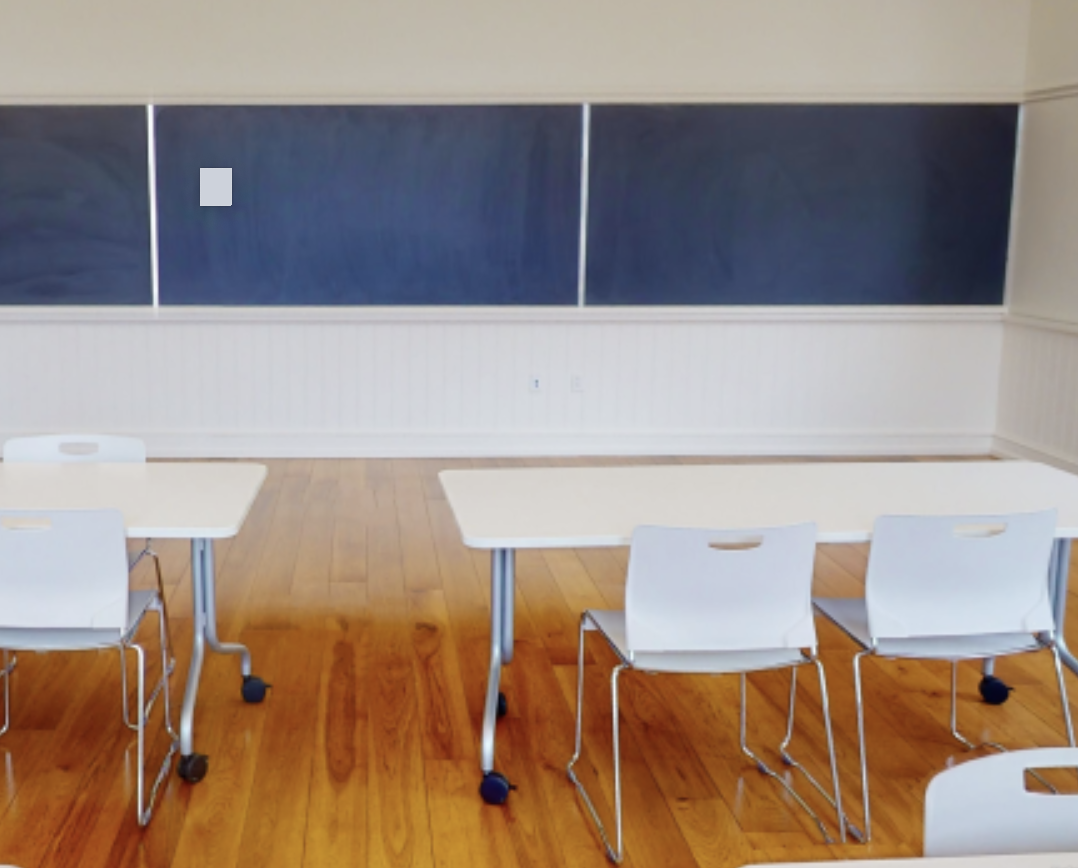} \\ 
    \footnotesize{(\textit{Second row view of the classroom with white tables facing a wall with blackboards.})}
    }\\
\midrule

\rowcolor{gray!15}
\multicolumn{4}{l}{\textbf{\correct}}\\
 Cogí un libro de un carrito pequeño.
& I picked up a book from a little cart.
& \parbox[p]{0.22\textwidth}{
    \includegraphics[alt={Image displaying small, low wooden shelf holding several books.}, width=0.22\textwidth]{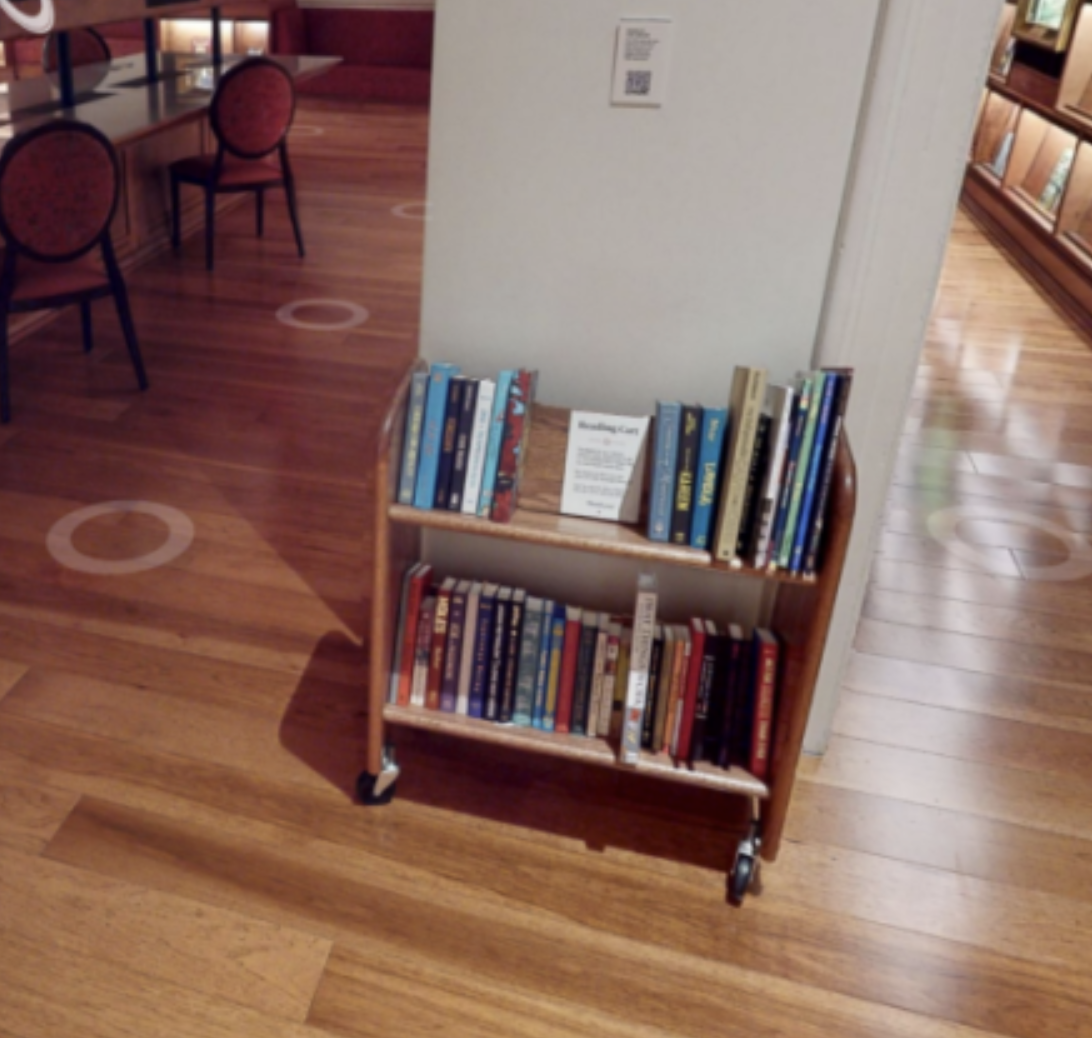} \\ 
    \footnotesize{(\textit{Small, low wooden shelf holding several books.})}
    }
& \parbox[p]{0.22\textwidth}{
    \includegraphics[alt={Image displaying small outdoor library box with doors that open to reveal books inside.}, width=0.22\textwidth]{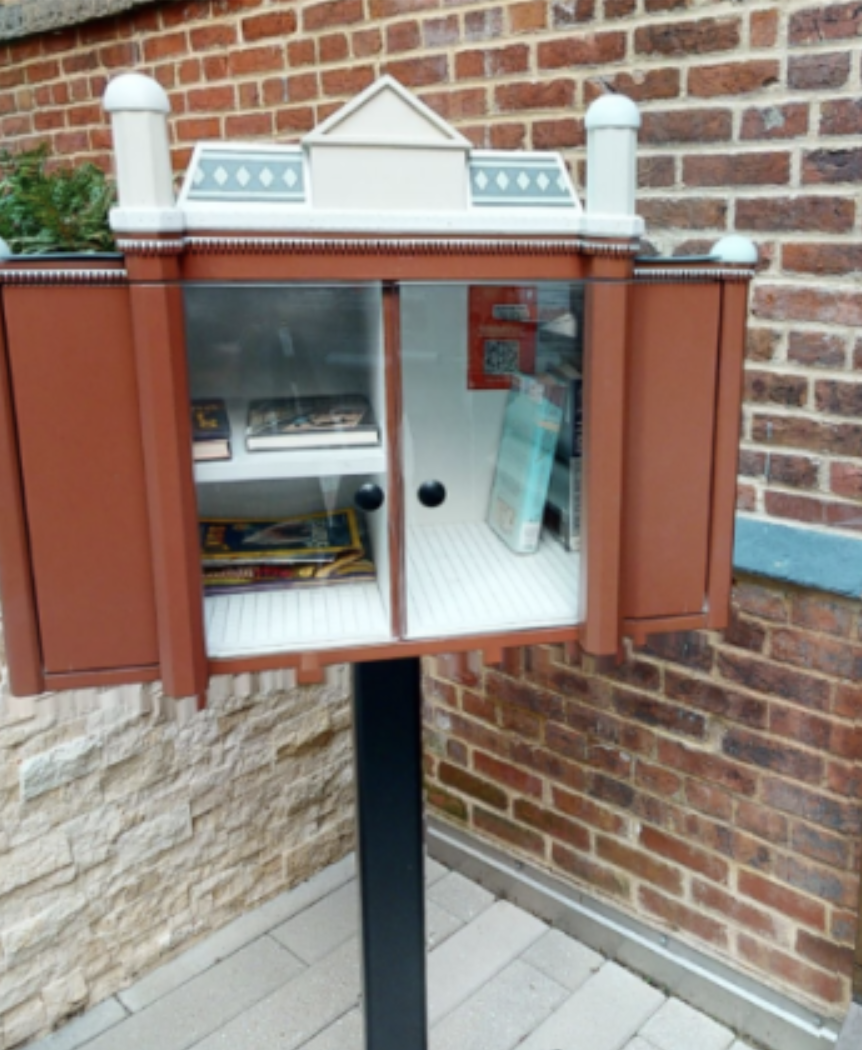} \\ 
    \footnotesize{(\textit{Small outdoor library box with doors that open to reveal books inside.})}
    }\\
\midrule

\rowcolor{gray!15}
\multicolumn{4}{l}{\textbf{\correct}}\\
 Casi me paso la puerta oculta.
& I almost missed the hidden door.
& \parbox[p]{0.22\textwidth}{
    \includegraphics[alt={Image displaying library with wooden shelves with a hidden door open.}, width=0.22\textwidth]{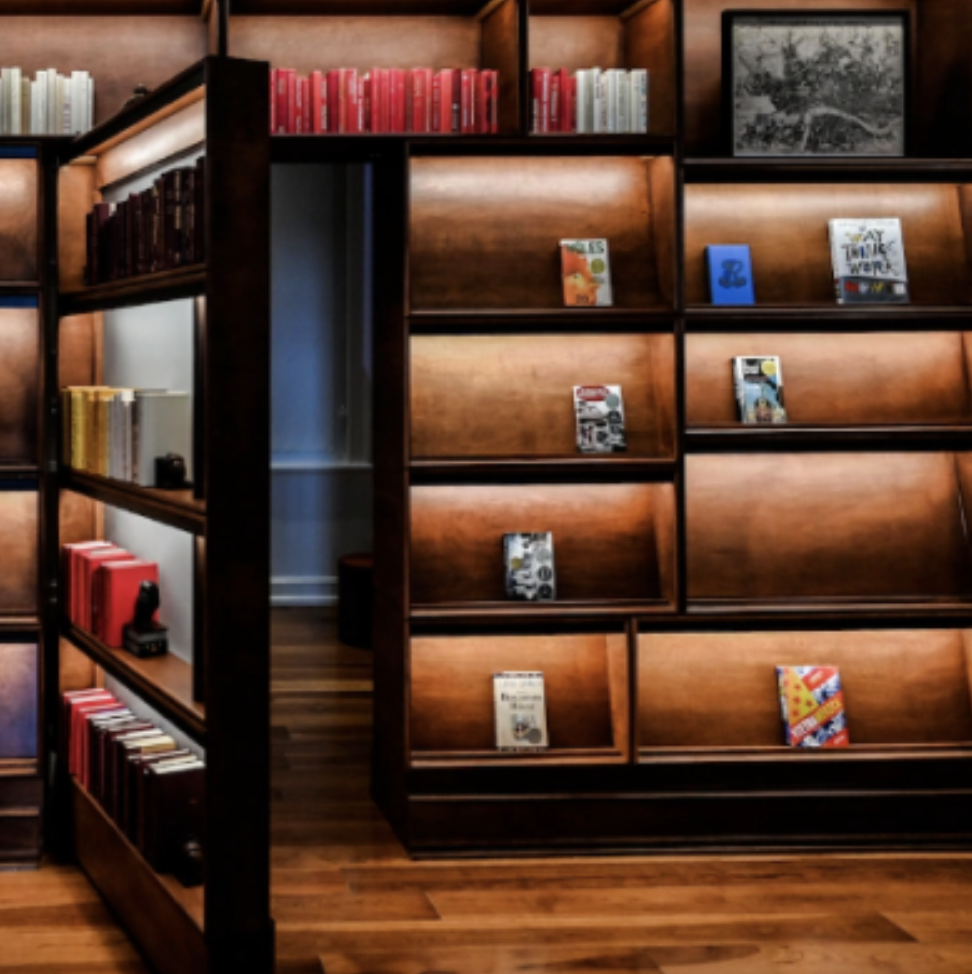} \\ 
    \footnotesize{(\textit{Library with wooden shelves with a hidden door open.})}
    }
& \parbox[p]{0.22\textwidth}{
    \includegraphics[alt={Image displaying doorway leading to another room with a wooden bookshelf.}, width=0.22\textwidth]{figs/appendix_imgs/s4_correct1_incorrect.png} \\ 
    \footnotesize{(\textit{Doorway leading to another room with a wooden bookshelf.})}
    }\\
 
\end{longtable*}